\documentclass[10pt,twocolumn,letterpaper]{article}

\usepackage[pagenumbers]{cvpr} % To force page numbers, e.g. for an arXiv version

\usepackage{booktabs}
\usepackage{multirow}
\usepackage{graphicx}
\usepackage[table]{xcolor}
\usepackage{tikz}
\usepackage{tabularx}
\usepackage{graphicx}
\usepackage{capt-of}
\usepackage{float}
\newsavebox{\probeablationbox}
\newsavebox{\qwenfigurebox}
\newsavebox{\qwenpanelbox}
\newsavebox{\ablationpanelbox}
\newlength{\pairedpanelheight}

\usepackage[most]{tcolorbox}
\usepackage{listings}

\definecolor{promptBlue}{HTML}{F4F8FC}
\definecolor{promptPurple}{HTML}{F8F5FC}
\definecolor{promptOrange}{HTML}{FCF7F1}
\definecolor{promptBorder}{HTML}{D5DCE5}

\lstdefinestyle{promptstyle}{
  basicstyle=\ttfamily\footnotesize,
  breaklines=true,
  breakatwhitespace=true,
  columns=fullflexible,
  keepspaces=true,
  showstringspaces=false,
  aboveskip=0pt,
  belowskip=0pt
}

\newtcblisting{promptbox}[2]{
  enhanced,
  breakable,
  listing only,
  listing options={
    style=promptstyle,
    breaklines=true,
    breakautoindent=false,
    breakindent=0pt,
    xleftmargin=0pt,
    xrightmargin=0pt,
    columns=fullflexible,
    keepspaces=true,
    showstringspaces=false
  },
  title={#1},
  colback=#2,
  colbacktitle=#2,
  colframe=promptBorder,
  coltitle=black,
  fonttitle=\small\bfseries,
  boxrule=0.5pt,
  arc=2pt,
  left=5pt,
  right=5pt,
  top=5pt,
  bottom=5pt,
  before skip=6pt,
  after skip=6pt
}

\usepackage{algorithm}
\usepackage{algpseudocode}

\definecolor{cvprblue}{rgb}{0.21,0.49,0.74}
\usepackage[pagebackref,breaklinks,colorlinks,allcolors=cvprblue]{hyperref}

\def\confName{CVPR}
\def\confYear{2027}

\definecolor{ssQwen}{HTML}{615CED}
\definecolor{ssTarsAqua}{RGB}{120,230,221}
\definecolor{ssTarsCyan}{RGB}{0,199,210}
\definecolor{ssTarsBlue}{RGB}{60,140,255}
\definecolor{ssTarsNavy}{RGB}{50,89,181}

\newcommand{\QwenIcon}{%
  \raisebox{-0.15em}{%
    \begin{tikzpicture}[x=0.042em,y=-0.042em]
      \path[use as bounding box] (0,0) rectangle (24,24);
      \fill[ssQwen]
        (19.713000,2.825000)
        -- (13.806000,2.825000)
        .. controls (13.806000,2.574000) and (13.740000,2.322000) .. (13.610000,2.097000)
        -- (12.819000,0.728000)
        .. controls (12.558477,0.278402) and (12.078625,0.001154) .. (11.559000,-0.000000)
        -- (9.979000,-0.000000)
        .. controls (9.459000,-0.000000) and (8.979000,0.275000) .. (8.719000,0.728000)
        -- (6.299000,4.918000)
        .. controls (6.038826,5.367393) and (6.038826,5.921607) .. (6.299000,6.371000)
        -- (7.090000,7.741000)
        .. controls (7.350000,8.189000) and (7.830000,8.468000) .. (8.350000,8.468000)
        -- (19.710000,8.468000)
        .. controls (20.230000,8.468000) and (20.710000,8.193000) .. (20.970000,7.740000)
        -- (21.760000,6.370000)
        .. controls (22.019403,5.921114) and (22.019403,5.367886) .. (21.760000,4.919000)
        -- (20.970000,3.549000)
        .. controls (20.709477,3.099402) and (20.229625,2.822154) .. (19.710000,2.821000)
        -- cycle
        (0.195000,9.905000)
        -- (3.146000,15.020000)
        .. controls (2.925729,15.147674) and (2.742674,15.330729) .. (2.615000,15.551000)
        -- (1.824000,16.921000)
        .. controls (1.564212,17.370140) and (1.564212,17.923860) .. (1.824000,18.373000)
        -- (2.614000,19.743000)
        .. controls (2.874000,20.191000) and (3.354000,20.470000) .. (3.874000,20.470000)
        -- (8.714000,20.470000)
        .. controls (9.234000,20.470000) and (9.714000,20.195000) .. (9.974000,19.742000)
        -- (10.764000,18.372000)
        .. controls (11.023403,17.923114) and (11.023403,17.369886) .. (10.764000,16.921000)
        -- (5.086000,7.083000)
        .. controls (4.825477,6.633402) and (4.345625,6.356154) .. (3.826000,6.355000)
        -- (2.246000,6.355000)
        .. controls (1.726000,6.355000) and (1.246000,6.631000) .. (0.986000,7.083000)
        -- (0.196000,8.453000)
        .. controls (-0.063788,8.902140) and (-0.063788,9.455860) .. (0.196000,9.905000)
        (16.084000,23.272000)
        -- (19.035000,18.157000)
        .. controls (19.252000,18.282000) and (19.504000,18.353000) .. (19.763000,18.353000)
        -- (21.345000,18.353000)
        .. controls (21.865000,18.353000) and (22.345000,18.078000) .. (22.605000,17.625000)
        -- (23.395000,16.255000)
        .. controls (23.654403,15.806114) and (23.654403,15.252886) .. (23.395000,14.804000)
        -- (20.975000,10.613000)
        .. controls (20.714656,10.163710) and (20.235267,9.886509) .. (19.716000,9.885000)
        -- (18.134000,9.885000)
        .. controls (17.614000,9.885000) and (17.134000,10.161000) .. (16.875000,10.613000)
        -- (11.197000,20.451000)
        .. controls (10.937212,20.900140) and (10.937212,21.453860) .. (11.197000,21.903000)
        -- (11.987000,23.273000)
        .. controls (12.247000,23.721000) and (12.727000,24.000000) .. (13.247000,24.000000)
        -- (14.829000,24.000000)
        .. controls (15.349000,24.000000) and (15.829000,23.725000) .. (16.089000,23.272000)
        -- cycle;
    \end{tikzpicture}%
  }%
}
\newcommand{\UITARSIcon}{%
  \raisebox{-0.15em}{%
    \begin{tikzpicture}[x=0.00265em,y=-0.00265em]
      \path[use as bounding box] (0,0) rectangle (381,364);
      \fill[ssTarsAqua] (20,12) -- (79,12) -- (220,350) -- (160,350) -- cycle;
      \fill[ssTarsCyan] (10,350) -- (256,12) -- (316,12) -- (69,350) -- cycle;
      \fill[ssTarsBlue] (69,350) -- (128,350) -- (375,12) -- (316,12) -- cycle;
      \fill[ssTarsNavy] (148,12) -- (207,12) -- (348,350) -- (288,350) -- cycle;
    \end{tikzpicture}%
  }%
}

\newcommand{\QwenBase}[1]{\QwenIcon\,#1}
\newcommand{\UITARSBase}[1]{\UITARSIcon\,#1}

\title{Your VLM Knows What to Inspect:\\ Semantic Probe-Guided Adaptive Focusing for GUI Grounding}

\author{%
  Ruixiao Shi\textsuperscript{1,2}\quad
  Fu Feng\textsuperscript{1,2}\quad
  Yucheng Xie\textsuperscript{1,2}\quad
  Jing Wang\textsuperscript{1,2*}\quad
  Xin Geng\textsuperscript{1,2*}\\
  \textsuperscript{1}School of Computer Science and Engineering,
  Southeast University, Nanjing, China\\
  \textsuperscript{2}Key Laboratory of New Generation Artificial
  Intelligence Technology and Its Interdisciplinary\\
  Applications (Southeast University), Ministry of Education, China\\[3pt]
  {\small\ttfamily
    \{eric\_xiao, fufeng, xieyc,  wangjing91, xgeng\}@seu.edu.cn}%
}

\begin{document}
\twocolumn[{
\maketitle
\begin{figure}[H]
\hsize=\textwidth 
\centering
\vspace{-0.2in}
\includegraphics[width=1\textwidth]{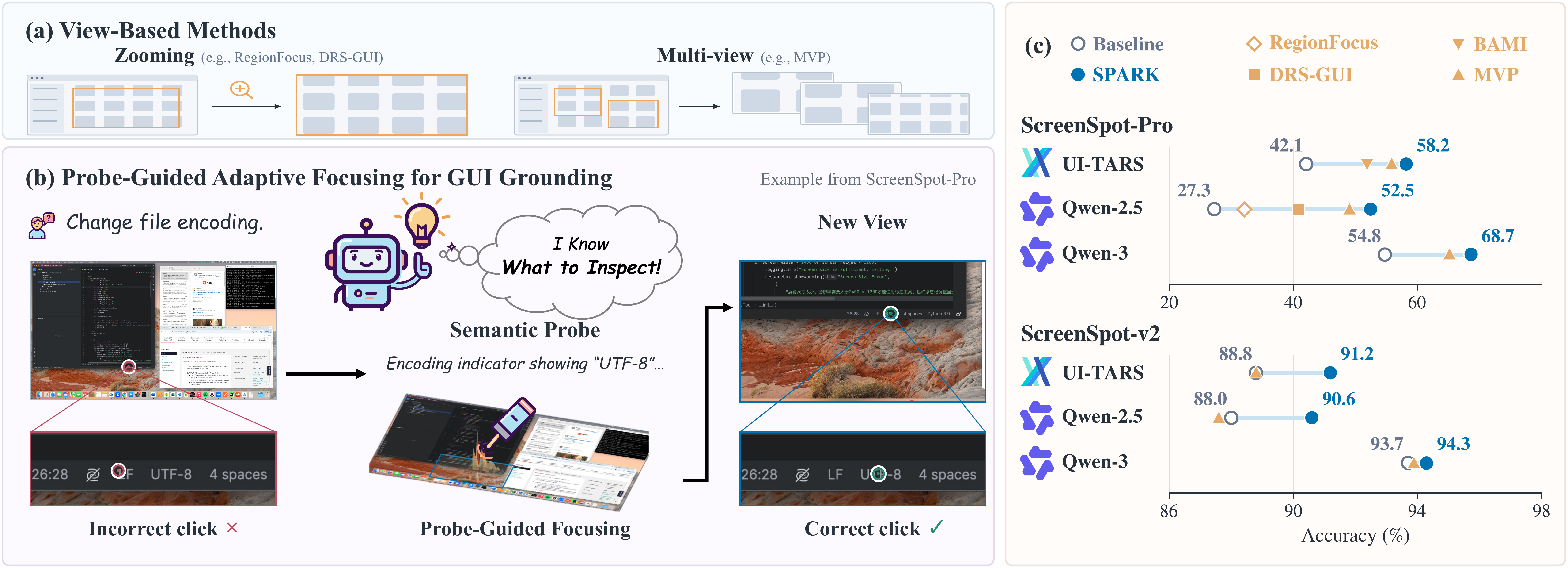}
\vspace{-0.25in}
\caption{\textbf{Your VLM knows what to inspect.}
    (a) Existing view-based methods use zooming or multiple views to improve GUI grounding.
    (b) An incorrect coordinate prediction can coexist with useful semantic guidance: the VLM clicks the neighboring ``LF'' control, yet identifies ``UTF-8'' as the evidence to inspect.
    SPARK connects semantic probes with spatial attention to guide adaptive focusing, enabling a subsequent view to localize the target.
    (c) SPARK improves grounding accuracy across three backbones on ScreenSpot series benchmarks.
    }
\label{fig:intro}
\end{figure}
}]
\begingroup
  \renewcommand{\thefootnote}{*}
  \footnotetext{Corresponding authors.}
\endgroup

\begin{abstract}
GUI grounding requires resolving fine visual details within the context that gives them meaning.
Additional views can improve localization, but their value depends on what evidence they expose and how much
context they preserve.
We explore a capability of vision-language models (VLMs): even when their coordinate predictions are incorrect, they can describe visual evidence worth inspecting.
We introduce \textbf{SPARK}, \textit{\textbf{S}emantic \textbf{P}robe-guided \textbf{A}daptive \textbf{R}egion see\textbf{K}ing},
a training-free framework that turns this semantic guidance into targeted visual observations.
Exploration probes describe relevant context and distinguishing target details, while verification probes
identify evidence needed to assess emerging hypotheses.
Their spatial attention guides view placement, contextual responses determine crop extent, and observation history encourages complementary inspection.
The resulting predictions are combined through uncertainty-weighted voting.
On ScreenSpot-Pro, SPARK improves grounding accuracy over single-view inference by 13.9, 25.2, and 16.1 percentage points with Qwen3-VL, Qwen2.5-VL, and UI-TARS, outperforming all compared test-time methods for each backbone.
Consistent gains across four additional benchmarks demonstrate its effectiveness across diverse interfaces and grounding tasks.
These results establish semantic probing as an effective mechanism for turning a frozen VLM's understanding of what to inspect into more accurate GUI grounding.
\end{abstract}

% \section{Introduction}
\section{Introduction}
\label{sec:intro}
Interacting with graphical user interfaces (GUIs) requires agents to ground natural-language instructions in precise screen locations~\cite{nguyen2025gui, tang2025survey, Xie2024OSWorld, zhang2024large}.
Despite advances in vision-language models (VLMs), this task remains challenging in complex interfaces, where identifying a target requires resolving fine-grained visual distinctions within the appropriate functional context~\cite{seeclick, Hong2024CogAgent, Xu2025Aguvis}.
These requirements place competing demands on visual observation: a broad view preserves contextual relationships but can obscure small details, while a closer view reveals local appearance but may exclude the panel or landmark needed to interpret it~\cite{Wu2025GUIActor, Zhang2025MLLMsKnow}.
Effective GUI grounding therefore requires deciding not only where to act, but also where to look, at what scale, and which visual evidence to inspect.

To address these challenges, prior work has explored view-based methods (Fig.~\ref{fig:intro}(a)), including zooming~\cite{regionfocus, Zhang2026bami, drsgui} and multi-view inference~\cite{mvp} to expose visual details that are difficult to resolve from the full screenshot.
These approaches highlight the value of additional observations, whose effectiveness depends on both the regions selected and the context retained.
A central challenge is therefore to determine what evidence each new view should reveal: revisiting similar regions may add little information, while an overly tight crop may remove the context needed to distinguish plausible targets. 
Choosing an informative view thus requires understanding what remains to be resolved about the target.
This raises a natural question:

\par\smallskip
\noindent
\colorbox{blue!6}{%
  \parbox{\dimexpr\linewidth-2\fboxsep\relax}{%
    \centering\itshape
 Can a VLM itself identify what to inspect next \\ to improve GUI grounding?
  }%
}
\par\smallskip

The example in Fig.~\ref{fig:intro}(b) suggests that useful inspection guidance can emerge even before accurate localization.
Although the VLM initially predicts an incorrect location, it can still describe relevant context and distinguishing details that guide further inspection.
We elicit this capability through \emph{semantic probes}: short, task-conditioned descriptions of visual evidence to seek for localization.
Crucially, these probes provide both a semantic account of what to inspect and an attention-based spatial response indicating where that evidence may lie.
By extracting visual attention associated with the probe's descriptive tokens, we translate its language guidance into spatial evidence maps.
These maps guide observation placement and the extent of surrounding context, connecting \emph{what to inspect} with \emph{where and at what scale to look}.

Building on this idea, we propose \textbf{SPARK}, \textit{\textbf{S}emantic \textbf{P}robe-guided
\textbf{A}daptive \textbf{R}egion see\textbf{K}ing},
a training-free framework that uses a frozen VLM's semantic guidance to construct a limited set of
local observations.
SPARK represents inspection guidance through
\emph{hierarchical semantic probes}:
\emph{Context} identifies relevant regions and landmarks, while \emph{Detail} specifies the visual evidence needed to identify or verify the requested control.
This shared structure supports two complementary forms of inspection.
An \emph{Exploration Probe}, conditioned on the instruction, describes evidence to seek;
a \emph{Verification Probe}, additionally conditioned on emerging localization hypotheses, describes evidence to inspect when assessing those hypotheses.
The spatial attention associated with these descriptions jointly guides view placement, while the Context response determines how much surrounding evidence to retain.
A shared adaptive focusing mechanism translates this guidance into crop locations and scales, using observation history to encourage spatial diversity.
Finally, \emph{uncertainty-weighted voting} integrates the resulting coordinate predictions into the final grounding decision.

We evaluate SPARK on five GUI grounding benchmarks using general-purpose and GUI-specialized VLMs.
On ScreenSpot-Pro, SPARK improves accuracy over single-view inference by an average of 18.4 percentage points (pp) across the three backbones, with gains of up to 25.2 pp. 
It also delivers consistent improvements across the remaining four benchmarks.

Our contributions are as follows: (1) We introduce \textit{semantic probes} to elicit a frozen VLM's latent inspection capability, connecting descriptions of task-relevant evidence with their spatial attention to guide where and at what scale to observe. (2) We propose \textbf{SPARK}, a training-free framework that unifies exploration and verification through hierarchical semantic probes, adaptively guiding where to inspect and how much context to retain within a limited local-view budget. (3) We demonstrate consistent improvements of SPARK across five GUI grounding benchmarks and four VLM backbones.

\section{Related Work}
\paragraph{GUI Grounding.}
GUI grounding maps natural-language instructions to actionable screen locations~\cite{Pasupat2018Mapping, seeclick, Li2020Mapping}.
SeeClick~\cite{seeclick}, OS-Atlas~\cite{osatlas}, and UGround-V1~\cite{uground} develop this capability through GUI-specific supervision.
UI-TARS~\cite{uitars} integrates perception, action modeling, and reasoning, while UI-TARS-1.5~\cite{uitars15} and GTA1~\cite{yang2026gta1} further improve GUI capabilities through reinforcement learning.

Building on these grounding capabilities, \textit{\textbf{test-time methods}} improve localization by exploiting pretrained models without additional task-specific training.
TAG~\cite{tag} and Trifuse~\cite{trifuse} exploit pretrained attention and complementary spatial cues.
RegionFocus~\cite{regionfocus} and DiMo-GUI~\cite{dimogui} use zooming, MVP~\cite{mvp} aggregates attention-guided views, DRS-GUI~\cite{drsgui} schedules perceptual actions, and BAMI~\cite{Zhang2026bami} combines coarse-to-fine focusing with candidate selection.
SPARK instead uses semantic inspection guidance to organize adaptive observations for exploration and verification.

\vspace{-0.1in}
\paragraph{Active Visual Perception and Reasoning.}
Active visual perception uses task understanding to guide the acquisition of relevant visual evidence.
V*~\cite{vstar} and ZoomEye~\cite{Shen2025ZoomEye} enable visual search, Visual CoT~\cite{visualcot} supports region localization, VisRL~\cite{visrl} learns to focus, and ViCrop~\cite{Zhang2025MLLMsKnow} uses internal model signals for training-free cropping.
SPARK elicits hierarchical descriptions from a frozen VLM and translates their spatial attention into task- and hypothesis-conditioned observations for GUI grounding.
\section{Method}
\label{sec:method}

\begin{figure*}[t]
  \centering
 
   \includegraphics[width=\linewidth]{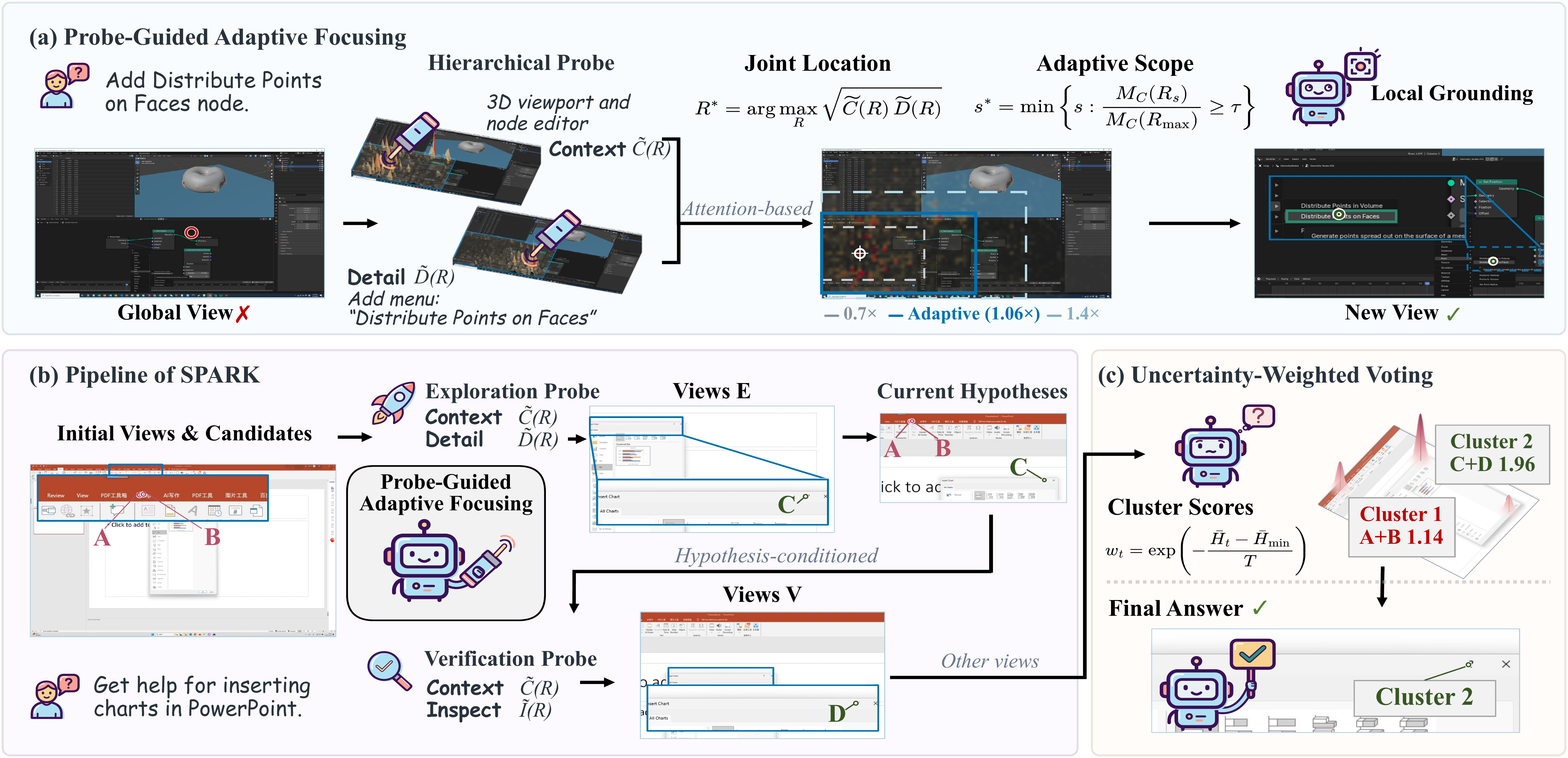}
    \vspace{-0.26in}
    \caption{
    \textbf{Overview of SPARK.}
    (a) Probe-guided adaptive focusing translates contextual and local-detail attention into view placement and context-dependent crop extent.
    (b) SPARK uses the focusing mechanism in (a) to construct local views guided by exploration and hypothesis-conditioned verification probes.
    The resulting localization predictions are passed to the aggregation module.
    (c) Uncertainty-weighted voting accumulates entropy-based prediction weights within spatial clusters and returns the medoid of the highest-scoring cluster as the final location.
    }
    \vspace{-0.05in}
    \label{fig:twocol}
\end{figure*}

\subsection{Preliminaries}
\label{sec:preliminaries}

\paragraph{View-based GUI grounding.}
Given a screenshot $I$ and a natural-language instruction $d$, GUI grounding predicts a target
coordinate in the original image space.
View-based methods gather additional visual evidence by constructing views at different locations and scales.
For a view region $R_t$, we write
\begin{equation}
    I_t = \mathcal{T}(I[R_t]), \qquad
    y_t = f_\theta(I_t,d), \qquad
    p_t = \phi_{R_t}(y_t),
\end{equation}
where $f_\theta$ is a frozen VLM, $\mathcal{T}$ preprocesses the cropped image,
and $\phi_{R_t}$ parses the generated coordinate and maps it to the original screenshot.
The initial region $R_0$ covers the full image.
A local-view budget $B$ limits the number of additional views, each subject to a visual token budget $N$.

View construction determines which visual evidence the model receives.
SPARK guides this process through \textit{semantic probes}, using the model's descriptions and their spatial attention to determine the location and extent of subsequent views.

\subsection{Probe-Guided Adaptive Focusing}
\label{sec:adaptive_focusing}

SPARK organizes view construction around a hierarchical
representation of inspection requirements.
Different conditioning signals determine the purpose
of inspection, while a shared attention-based mechanism
translates these requirements into adaptive local views.

\vspace{-0.05in}
\paragraph{Hierarchical semantic probes.}
A \emph{semantic probe} is a short, task-conditioned description of visual evidence to inspect for localization.
We organize this description at two complementary levels:
\emph{Context} identifies a relevant region and its landmarks,
while \emph{Detail} specifies the local evidence needed to recognize or assess the requested control.
This hierarchy connects the surrounding context that gives an element meaning with the visual features that distinguish it.

We instantiate this representation through two forms of conditioning.
The \emph{Exploration Probe} uses the screenshot and instruction to describe the target's context and distinguishing details:
\begin{equation}
    (z_C^E,z_D^E)
    =
    f_\theta(I,d;\pi_E),
    \label{eq:exploration_probe}
\end{equation}
where $\pi_E$ is the exploration prompt. 
The Context description $z_C^E$ identifies the likely containing panel or screen region and relevant landmarks.
The Detail description $z_D^E$ identifies the requested control and features that distinguish it from nearby elements.
Together, they express where to seek the target and what evidence to look for.

The \emph{Verification Probe} additionally conditions inspection on localization hypotheses formed from previous views.
Let $\mathcal{P}^{\mathrm{hyp}}_t$ contain the medoids of the two highest-scoring prediction clusters under Sec.~\ref{sec:voting}, or the single medoid when only one cluster is available.
We mark these locations on the screenshot to obtain $I^{\mathcal{P}^{\mathrm{hyp}}_t}$ and generate
\begin{equation}
    (z_{\mathrm{ctrl}}^V,z_C^V,z_I^V)
    =
    f_\theta
    \bigl(I^{\mathcal{P}^{\mathrm{hyp}}_t},d;\pi_V\bigr),
    \label{eq:verification_probe}
\end{equation}
where $\pi_V$ asks for evidence needed to assess the candidate locations.
The Control field $z_{\mathrm{ctrl}}^V$ specifies the requested target.
The Context field $z_C^V$ describes its relevant surroundings, while the Inspect field $z_I^V$ provides detail-level guidance by identifying a concrete feature or landmark needed to check its identity or function.
Context and Inspect provide the spatial guidance used for verification.
Candidate labels follow spatial order.

Exploration seeks evidence relevant to the task, whereas verification seeks evidence that can support or revise the current hypotheses.
Both express inspection requirements through the \textit{context-to-detail hierarchy}, allowing semantic guidance to evolve with the model's observations.

\vspace{-0.05in}
\paragraph{From semantic probes to spatial guidance.}
The attention associated with probe content tokens connects semantic inspection requirements to image regions.
For a description $z$, let $\mathcal{Q}(z)$ denote its content-query positions.
For visual token $i$, we compute
\begin{equation}
    A_z(i)=
    \frac{1}{|\mathcal{Q}(z)|}
    \sum_{q\in\mathcal{Q}(z)}
    \max_h a^{(\ell,h)}_{q,i},
    \quad
    \bar A_z(i)=\frac{A_z(i)}{\sum_j A_z(j)}.
    \label{eq:probe_attention}
\end{equation}
where $a^{(\ell,h)}_{q,i}$ is the softmax attention weight at layer $\ell$ and head $h$.
The softmax includes all visible keys before we extract the visual entries.
Taking the per-patch maximum across heads retains spatial responses expressed by different heads;
averaging across content queries combines the evidence associated with the description.
Visual token centers $x_i$ are mapped to original-image coordinates using the actual preprocessing grid.

Each probe yields a pair of normalized spatial maps:
\begin{equation}
    (C_E,D_E)=(\bar A_{z_C^E},\bar A_{z_D^E}),
    \quad
    (C_V,I_V)=(\bar A_{z_C^V},\bar A_{z_I^V}).
\end{equation}
The two maps represent complementary aspects of the same inspection requirement:
contextual support and local visual evidence.

We construct candidate windows from the top-$K$ locations in $C_E$, $D_E$, and the initial global localization attention $A_G$.
Each location anchors a standard window adjusted to the image boundaries, and identical rectangles
are removed.
Exploration and verification prioritize this shared candidate pool $\mathcal{P}$ through their respective map pairs.
The probe's semantic purpose therefore determines which available region is most relevant to inspect.

\vspace{-0.05in}
\paragraph{Adaptive view construction.}
SPARK jointly uses contextual and local evidence to select a view and adapts its extent to the
spatial distribution of contextual support.
View history further promotes spatial diversity.
Let $\mathcal{H}_t$ contain the previously executed local regions.
We discount evidence already covered by these views:
\begin{equation}
    \omega_t(i)
    =
    1-\rho\,
    \mathbf{1}\left[
        x_i\in\bigcup_{R\in\mathcal{H}_t}R
    \right],
    \label{eq:history_discount}
\end{equation}
where $\rho$ controls the strength of the coverage discount.

For each candidate window $R$, we aggregate the history-discounted attention within the region.
For exploration, the Context and Detail responses are
\begin{equation}
    \begin{aligned}
        \widetilde C_t^E(R)
        &= \sum_{i:x_i\in R}\omega_t(i)C_E(i),
        \\
        \widetilde D_t^E(R)
        &= \sum_{i:x_i\in R}\omega_t(i)D_E(i).
    \end{aligned}
\end{equation}
The verification responses $\widetilde C_t^V(R)$ and $\widetilde I_t^V(R)$ are computed analogously from $C_V$ and $I_V$.

Both probe types score candidate windows through joint contextual and local evidence:
\begin{equation}
    S_t(R)=
    \begin{cases}
        \sqrt{\widetilde C_t^E(R)\,
              \widetilde D_t^E(R)},
        & \text{for exploration}, \\[3pt]
        \sqrt{\widetilde C_t^V(R)\,
              \widetilde I_t^V(R)},
        & \text{for verification}.
    \end{cases}
    \label{eq:joint_region_score}
\end{equation}
We select the standard candidate window
\begin{equation}
    R_t^\ast
    =
    \arg\max_{R\in\mathcal P} S_t(R).
    \label{eq:joint_region_selection}
\end{equation}
The geometric mean favors windows supported by both contextual and local evidence.
History discounting encourages complementary views while allowing a relevant region to be revisited.
% Figure~\ref{fig:twocol} denotes these window responses by $\widetilde C(R)$, $\widetilde D(R)$, and $\widetilde I(R)$, omitting the probe-type and step indices.

After selecting $R_t^\ast$, we retain its associated anchor $c$ and adapt the crop size to the spatial extent of contextual evidence.
The exploration Context map $C_E$ serves as a persistent reference for scale selection in
both exploration and verification.
For standard window dimensions $w_0$ and $h_0$, we define the locally weighted response
{\small
\begin{equation}
    C_c^{\mathrm{loc}}(i)
    =
    C_E(i)
    \exp\!\left[
        -\frac{1}{2}
        \left(
            \frac{(x_{i,x}-c_x)^2}{(w_0/2)^2}
            +
            \frac{(x_{i,y}-c_y)^2}{(h_0/2)^2}
        \right)
    \right].
    \label{eq:local_context}
\end{equation}
}
Let $R_s=R(c,s)$ denote the boundary-adjusted window at scale $s$, and let $R_{\max}=R(c,s_{\max})$ denote the maximum reference window.
For the selected anchor, the retained contextual mass is
\begin{equation}
    M_C(R_s)
    =
    \sum_{i:x_i\in R_s}C_c^{\mathrm{loc}}(i).
\end{equation}
We choose the smallest scale that retains a fraction $\tau$ of the contextual mass within
$R_{\max}$:
\begin{equation}
    s^\ast
    =
    \min\left\{
        s\in[s_{\min},s_{\max}]:
        \frac{M_C(R_s)}{M_C(R_{\max})}
        \geq\tau
    \right\}.
    \label{eq:adaptive_scale}
\end{equation}
We use $[s_{\min},s_{\max}]=[0.7,1.4]$ and $\tau=0.9$.
Concentrated evidence produces a tighter view, while distributed contextual support expands
the observation scope.
This scale decision uses the original Context response without history discounting, preserving the evidence needed to interpret each view.

The final observation region is $R_t=R(c,s^\ast)$.
Its crop is processed within the visual token budget $N$, and $R_t$ is added to the view history:
\begin{equation}
    \mathcal{H}_{t+1}
    =
    \mathcal{H}_t\cup\{R_t\}.
\end{equation}

Following prior work~\cite{mvp}, we initialize local inspection from the global localization attention, with this initial local view counted toward the view budget.
Subsequent views are constructed through hierarchical semantic probes: after each exploration step, competing localization hypotheses trigger hypothesis-conditioned verification.
The resulting predictions are integrated by the following voting rule.

\subsection{Uncertainty-Weighted Voting}
\label{sec:voting}

Complementary observations can produce competing localization hypotheses.
SPARK combines their spatial agreement with predictive uncertainty, allowing confident evidence to receive greater influence in the final decision.

For localization output $y_t$, let $\mathcal{K}_t$ contain its generated token positions, excluding special and purely whitespace tokens.
We measure uncertainty using the mean predictive entropy:
\begin{equation}
    \bar H_t
    =
    -\frac{1}{|\mathcal{K}_t|}
    \sum_{k\in\mathcal{K}_t}
    \sum_{v\in\mathcal{V}}
    P_{t,k}(v)\log P_{t,k}(v),
    \label{eq:output_entropy}
\end{equation}
where $\mathcal{V}$ is the vocabulary and $P_{t,k}$ is the next-token distribution conditioned on the view, instruction, and preceding output.

We group valid original-image predictions using complete-link clustering with threshold $\epsilon$.
Predictions are processed in observation order, and a point joins a cluster when its $\ell_\infty$ distance to every member is at most $\epsilon$.
Each prediction receives an uncertainty-dependent weight:
\begin{equation}
    w_t
    =
    \exp\left(
        -\frac{\bar H_t-\bar H_{\min}}{T}
    \right),
    \qquad
    \bar H_{\min}=\min_j\bar H_j.
\end{equation}
The winning cluster is
\begin{equation}
    \mathcal{C}^\ast
    =
    \arg\max_{\mathcal{C}}
    \sum_{t\in\mathcal{C}}w_t.
    \label{eq:weighted_vote}
\end{equation}
This rule combines support across observations while allowing a confident prediction to outweigh several uncertain alternatives.
Global and local predictions participate in the same aggregation.

Within the winning cluster, the final coordinate is its spatial medoid:
\begin{equation}
    t^\ast
    =
    \arg\min_{t\in\mathcal{C}^\ast}
    \sum_{j\in\mathcal{C}^\ast}
    \|p_t-p_j\|_2,
    \qquad
    p^\ast=p_{t^\ast}.
\end{equation}
The medoid preserves an observed prediction and represents the spatial consensus of the selected hypothesis.
We use $\epsilon=14$ pixels and $T=0.3$ by default.

\begin{table*}[!t]
  \centering
  \caption{Grounding accuracy (\%) on ScreenSpot-Pro (SSPro). Best overall accuracy per backbone is shown in \textbf{bold}.}
  \vspace{-0.05in}
  \label{tab:sspro}
  \small
  \setlength{\tabcolsep}{2pt}
  \renewcommand{\arraystretch}{1.12}
  \resizebox{\textwidth}{!}{%
    \begin{tabular}{@{}ll
      >{\columncolor{gray!5}\centering\arraybackslash}m{2.1em}
      >{\columncolor{gray!5}\centering\arraybackslash}m{2.1em}
      >{\centering\arraybackslash}m{2.1em}
      >{\centering\arraybackslash}m{2.1em}
      >{\columncolor{gray!5}\centering\arraybackslash}m{2.1em}
      >{\columncolor{gray!5}\centering\arraybackslash}m{2.1em}
      >{\centering\arraybackslash}m{2.1em}
      >{\centering\arraybackslash}m{2.1em}
      >{\columncolor{gray!5}\centering\arraybackslash}m{2.1em}
      >{\columncolor{gray!5}\centering\arraybackslash}m{2.1em}
      >{\centering\arraybackslash}m{2.1em}
      >{\centering\arraybackslash}m{2.1em}c@{}}
      \toprule
      \multirow{2}{*}{\textbf{Model / Method}}
        & \multirow{2}{*}{\textbf{Base model}}
        & \multicolumn{2}{c}{\cellcolor{gray!12}\textbf{Dev.}}
        & \multicolumn{2}{c}{\textbf{Creative}}
        & \multicolumn{2}{c}{\cellcolor{gray!12}\textbf{CAD}}
        & \multicolumn{2}{c}{\textbf{Sci.}}
        & \multicolumn{2}{c}{\cellcolor{gray!12}\textbf{Office}}
        & \multicolumn{2}{c}{\textbf{OS}}
        & \multirow{2}{*}{\textbf{Avg.$\uparrow$}} \\
      \cmidrule(lr){3-4}\cmidrule(lr){5-6}\cmidrule(lr){7-8}
      \cmidrule(lr){9-10}\cmidrule(lr){11-12}\cmidrule(lr){13-14}
        & & \cellcolor{gray!12}Text & \cellcolor{gray!12}Icon
        & Text & Icon & \cellcolor{gray!12}Text & \cellcolor{gray!12}Icon
        & Text & Icon & \cellcolor{gray!12}Text & \cellcolor{gray!12}Icon
        & Text & Icon & \\
      \midrule
      \rowcolor{gray!18}
      \multicolumn{15}{@{}l@{}}{\hspace{3pt}\strut\textbf{General-purpose models}} \\
      GPT-4o~\cite{hurst2024gpt} & --
        & 1.3 & 0.0 & 1.0 & 0.0 & 2.0 & 0.0
        & 2.1 & 0.0 & 1.1 & 0.0 & 0.0 & 0.0 & 0.8 \\
      Claude Computer Use~\cite{hu2024dawn} & --
        & 22.0 & 3.9 & 25.9 & 3.4 & 14.5 & 3.7
        & 33.9 & 15.8 & 30.1 & 16.3 & 11.0 & 4.5 & 17.1 \\
      Qwen2.5-VL-7B-Instruct~\cite{qwen25vl} & --
        & 51.9 & 2.8 & 36.9 & 7.7 & 16.2 & 1.6
        & 50.7 & 7.3 & 53.1 & 18.9 & 36.4 & 7.9 & 27.3 \\
      Qwen3-VL-8B-Instruct~\cite{qwen3vl} & --
        & 78.6 & 24.8 & 69.7 & 18.9 & 58.9 & 14.1
        & 77.8 & 30.9 & 85.9 & 35.8 & 72.9 & 27.0 & 54.8 \\
      \midrule
      \rowcolor{gray!18}
      \multicolumn{15}{@{}l@{}}{\hspace{3pt}\strut\textbf{GUI-specific models}} \\
      OS-Atlas-7B~\cite{osatlas} & \QwenBase{2-VL-7B}
        & 33.1 & 1.4 & 28.8 & 2.8 & 12.2 & 4.7
        & 37.5 & 7.3 & 33.9 & 5.7 & 27.1 & 4.5 & 18.9 \\
      UGround-V1-7B~\cite{uground} & \QwenBase{2-VL-7B}
        & 51.9 & 2.8 & 47.5 & 9.8 & 18.8 & 1.6
        & 57.6 & 14.5 & 60.5 & 13.2 & 38.3 & 7.9 & 31.1 \\
      SE-GUI-7B~\cite{yuan2025enhancing} & \QwenBase{2.5-VL-7B}
        & 68.2 & 19.3 & 57.6 & 9.1 & 51.3 & 14.1
        & 75.0 & 28.2 & 78.5 & 43.4 & 49.5 & 25.8 & 47.2 \\
      GUI-G2-7B~\cite{tang2026gui} & \QwenBase{2.5-VL-7B}
        & 68.8 & 17.2 & 57.1 & 15.4 & 55.8 & 12.5
        & 77.1 & 24.5 & 74.0 & 32.7 & 57.9 & 21.3 & 47.5 \\
      UI-TARS-1.5-7B~\cite{uitars15} & \QwenBase{2.5-VL-7B}
        & 61.0 & 8.3 & 54.5 & 14.7 & 37.6 & 14.1
        & 70.8 & 24.5 & 77.4 & 41.5 & 45.8 & 11.2 & 42.1 \\
      \midrule
      \rowcolor{gray!18}
      \multicolumn{15}{@{}l@{}}{\hspace{3pt}\strut\textbf{Test-time methods}} \\
      \rowcolor{blue!6}
      \multicolumn{15}{@{}l@{}}{\hspace{3pt}\strut\textit{Backbone: }\UITARSIcon\ UI-TARS-1.5-7B} \\
      BAMI~\cite{Zhang2026bami} & \UITARSBase{1.5-7B}
        & 71.4 & 22.1 & 68.2 & 21.7 & 49.8 & 14.1
        & 77.8 & 23.6 & 82.5 & 41.5 & 69.1 & 24.2 & 51.9\,{\scriptsize\textcolor{teal!60!black}{(+9.8)}} \\
      MVP~\cite{mvp} & \UITARSBase{1.5-7B}
        & 71.4 & 28.3 & 67.2 & 27.3 & 64.5 & 18.8
        & 77.8 & 31.8 & 82.5 & 50.9 & 65.4 & 36.0 & 55.9\,{\scriptsize\textcolor{teal!60!black}{(+13.8)}} \\
      \rowcolor{violet!15}
      \textbf{SPARK} & \UITARSBase{1.5-7B}
        & 78.6 & 29.0 & 68.7 & 25.9 & 64.0 & 23.4
        & 77.8 & 31.8 & 84.7 & 54.7 & 74.8 & 41.6
        & \textbf{58.2}\,{\scriptsize\textcolor{teal!60!black}{(+16.1)}} \\
      \addlinespace[4pt]
      \cmidrule[0.35pt]{1-15}
      \rowcolor{blue!6}
      \multicolumn{15}{@{}l@{}}{\hspace{3pt}\strut\textit{Backbone: }\QwenIcon\ Qwen2.5-VL-7B} \\
      RegionFocus~\cite{regionfocus} & \QwenBase{2.5-VL-7B}
        & 53.2 & 3.4 & 42.9 & 4.9 & 28.4 & 3.1
        & 56.9 & 10.9 & 59.9 & 24.5 & 41.1 & 15.7 & 32.1\,{\scriptsize\textcolor{teal!60!black}{(+4.8)}} \\
      DRS-GUI~\cite{drsgui} & \QwenBase{2.5-VL-7B}
        & 60.4 & 13.1 & 54.6 & 16.8 & 42.1 & 9.4
        & 59.7 & 16.4 & 66.7 & 26.4 & 52.3 & 23.6 & 40.9\,{\scriptsize\textcolor{teal!60!black}{(+13.6)}} \\
      MVP~\cite{mvp} & \QwenBase{2.5-VL-7B}
        & 67.5 & 17.9 & 64.1 & 19.6 & 51.8 & 12.5
        & 72.2 & 22.7 & 80.2 & 32.1 & 61.7 & 31.5 & 49.1\,{\scriptsize\textcolor{teal!60!black}{(+21.8)}} \\
      \rowcolor{violet!15}
      \textbf{SPARK} & \QwenBase{2.5-VL-7B}
        & 74.7 & 22.1 & 66.7 & 20.3 & 50.3 & 14.1
        & 75.0 & 25.5 & 82.5 & 39.6 & 71.0 & 39.3 & \textbf{52.5}\,{\scriptsize\textcolor{teal!60!black}{(+25.2)}} \\
      \addlinespace[4pt]
      \cmidrule[0.35pt]{1-15}
      \rowcolor{blue!6}
      \multicolumn{15}{@{}l@{}}{\hspace{3pt}\strut\textit{Backbone: }\QwenIcon\ Qwen3-VL-8B} \\
      MVP~\cite{mvp} & \QwenBase{3-VL-8B}
        & 87.0 & 40.7 & 76.8 & 34.3 & 72.6 & 32.8
        & 83.3 & 38.2 & 89.3 & 56.6 & 78.5 & 43.8 & 65.2\,{\scriptsize\textcolor{teal!60!black}{(+10.4)}} \\
      \rowcolor{violet!15}
      \textbf{SPARK} & \QwenBase{3-VL-8B}
        & 87.0 & 39.3 & 79.3 & 37.8 & 75.1 & 40.6
        & 87.5 & 43.6 & 91.0 & 67.9 & 83.2 & 56.2 & \textbf{68.7}\,{\scriptsize\textcolor{teal!60!black}{(+13.9)}} \\
      \bottomrule
    \end{tabular}%
  }
\end{table*}

\section{Experimental Setup}
\label{sec:experiments}

\paragraph{Benchmarks, backbones and compared methods.}
We evaluate SPARK on five GUI grounding benchmarks:
ScreenSpot-\{Pro, v2, v1\}~\cite{li2025screenspot,osatlas,seeclick},
UI-Vision~\cite{uivision}, and OSWorld-G~\cite{osworldg}.
Our primary backbones are two general-purpose VLMs,
Qwen3-VL-8B~\cite{qwen3vl} and
Qwen2.5-VL-7B~\cite{qwen25vl},
and a GUI-specialized model,
UI-TARS-1.5-7B~\cite{uitars15}.
We additionally evaluate Qwen3-VL-32B~\cite{qwen3vl} to examine scalability to a larger backbone.
Our comparisons cover general-purpose VLMs,
GUI-specialized models, and test-time grounding methods.
The first two categories establish single-view baselines,
while the third includes RegionFocus~\cite{regionfocus}, MVP~\cite{mvp},
DRS-GUI~\cite{drsgui}, and BAMI~\cite{Zhang2026bami}.

\vspace{-0.05in}
\paragraph{Implementation details.}
All localization calls use greedy decoding and fixed backbone-specific prompts, with at most four local views per screenshot. For MVP, we adopted the same local-view budget to ensure comparability.
We extract attention at layer index 24 for Qwen3-VL-8B, 48 for Qwen3-VL-32B, and 20 for Qwen2.5-VL and UI-TARS, taking the per-patch maximum across heads and averaging over probe content queries.

\section{Results}

\subsection{Main Results}

\begin{table*}[!t]
  \centering
  \caption{Grounding accuracy (\%) on ScreenSpot-\{v1,v2\}, OSWorld-G, and UI-Vision. Best overall accuracy per backbone is shown in \textbf{bold}. N/A indicates that the Refusal (Ref.) category is not evaluated under our protocol, consistent with MVP~\cite{mvp}.}
  \vspace{-0.05in}
  \label{tab:grounding-four-benchmarks}
  
  \small
  \setlength{\tabcolsep}{2pt}
  \setlength{\extrarowheight}{1pt}
  \renewcommand{\tabularxcolumn}[1]{m{#1}}
  \begin{minipage}[t]{0.49\textwidth}
    \centering
    \textbf{(a) ScreenSpot-v1 (SSv1)}\\[3pt]
    \begin{tabularx}{\linewidth}{@{}l >{\hsize=0.85\hsize\linewidth=\hsize\columncolor{gray!5}\centering\arraybackslash}X >{\hsize=0.85\hsize\linewidth=\hsize\columncolor{gray!5}\centering\arraybackslash}X >{\hsize=0.85\hsize\linewidth=\hsize\centering\arraybackslash}X >{\hsize=0.85\hsize\linewidth=\hsize\centering\arraybackslash}X >{\hsize=0.85\hsize\linewidth=\hsize\columncolor{gray!5}\centering\arraybackslash}X >{\hsize=0.85\hsize\linewidth=\hsize\columncolor{gray!5}\centering\arraybackslash}X >{\hsize=1.9\hsize\linewidth=\hsize\centering\arraybackslash}X@{}}
      \toprule
      \multirow{2}{*}{\textbf{Model / Method}}
        & \multicolumn{2}{c}{\cellcolor{gray!12}Mobile}
        & \multicolumn{2}{c}{Desktop}
        & \multicolumn{2}{c}{\cellcolor{gray!12}Web}
        & \multirow{2}{*}{\textbf{Avg.}$\uparrow$} \\
      \cmidrule(lr){2-3}\cmidrule(lr){4-5}\cmidrule(lr){6-7}
        & Text & Icon & Text & Icon & Text & Icon & \\
      \midrule
      \rowcolor{blue!6}
      \UITARSIcon\ 1.5-7B & 90.8 & 78.6 & 92.8 & 75.7 & 87.0 & 84.5 & 85.5 \\
      \quad +MVP & 91.6 & 80.8 & 93.8 & 82.1 & 85.7 & 78.6 & 85.8\,{\scriptsize\textcolor{teal!60!black}{(+0.3)}} \\
      \rowcolor{violet!15}
      \quad +\textbf{SPARK} & 94.1 & 83.0 & 94.3 & 84.3 & 89.1 & 85.0 & \textbf{88.7}\,{\scriptsize\textcolor{teal!60!black}{(+3.2)}} \\
      \addlinespace[3pt]
      \cmidrule[0.35pt]{1-8}
      \rowcolor{blue!6}
      \QwenIcon\ 2.5-VL-7B & 97.1 & 76.9 & 88.7 & 67.9 & 89.1 & 80.1 & 84.7 \\
      \quad +MVP & 90.8 & 77.7 & 89.2 & 74.3 & 86.1 & 79.1 & 83.6\,{\scriptsize\textcolor{red!65!black}{(-1.1)}} \\
      \rowcolor{violet!15}
      \quad +\textbf{SPARK} & 94.1 & 80.3 & 93.3 & 81.4 & 90.4 & 82.5 & \textbf{87.6}\,{\scriptsize\textcolor{teal!60!black}{(+2.9)}} \\
      \addlinespace[3pt]
      \cmidrule[0.35pt]{1-8}
      \rowcolor{blue!6}
      \QwenIcon\ 3-VL-8B & 97.1 & 80.8 & 96.4 & 85.0 & 92.2 & 84.5 & 89.8 \\
      \quad +MVP & 98.2 & 79.5 & 96.4 & 90.7 & 91.3 & 84.0 & 90.2\,{\scriptsize\textcolor{teal!60!black}{(+0.4)}} \\
      \rowcolor{violet!15}
      \quad +\textbf{SPARK} & 96.7 & 82.1 & 96.4 & 91.4 & 94.3 & 85.0 & \textbf{91.1}\,{\scriptsize\textcolor{teal!60!black}{(+1.3)}} \\
      \bottomrule
    \end{tabularx}
  \end{minipage}\hfill%
  \begin{minipage}[t]{0.49\textwidth}
    \centering
    \textbf{(b) ScreenSpot-v2 (SSv2)}\\[3pt]
    \begin{tabularx}{\linewidth}{@{}l >{\hsize=0.85\hsize\linewidth=\hsize\columncolor{gray!5}\centering\arraybackslash}X >{\hsize=0.85\hsize\linewidth=\hsize\columncolor{gray!5}\centering\arraybackslash}X >{\hsize=0.85\hsize\linewidth=\hsize\centering\arraybackslash}X >{\hsize=0.85\hsize\linewidth=\hsize\centering\arraybackslash}X >{\hsize=0.85\hsize\linewidth=\hsize\columncolor{gray!5}\centering\arraybackslash}X >{\hsize=0.85\hsize\linewidth=\hsize\columncolor{gray!5}\centering\arraybackslash}X >{\hsize=1.9\hsize\linewidth=\hsize\centering\arraybackslash}X@{}}
      \toprule
      \multirow{2}{*}{\textbf{Model / Method}}
        & \multicolumn{2}{c}{\cellcolor{gray!12}Mobile}
        & \multicolumn{2}{c}{Desktop}
        & \multicolumn{2}{c}{\cellcolor{gray!12}Web}
        & \multirow{2}{*}{\textbf{Avg.}$\uparrow$} \\
      \cmidrule(lr){2-3}\cmidrule(lr){4-5}\cmidrule(lr){6-7}
        & Text & Icon & Text & Icon & Text & Icon & \\
      \midrule
      \rowcolor{blue!6}
      \UITARSIcon\ 1.5-7B & 93.4 & 82.9 & 95.4 & 78.6 & 90.2 & 87.7 & 88.8 \\
      \quad +MVP & 95.2 & 82.5 & 96.9 & 83.6 & 89.3 & 81.8 & 88.8\,{\scriptsize(0.0)} \\
      \rowcolor{violet!15}
      \quad +\textbf{SPARK} & 95.9 & 86.3 & 96.9 & 85.7 & 92.7 & 86.2 & \textbf{91.2}\,{\scriptsize\textcolor{teal!60!black}{(+2.4)}} \\
      \addlinespace[3pt]
      \cmidrule[0.35pt]{1-8}
      \rowcolor{blue!6}
      \QwenIcon\ 2.5-VL-7B & 98.6 & 84.8 & 90.7 & 70.0 & 92.3 & 80.8 & 88.0 \\
      \quad +MVP & 91.7 & 85.8 & 92.3 & 78.6 & 91.0 & 81.3 & 87.6\,{\scriptsize\textcolor{red!65!black}{(-0.4)}} \\
      \rowcolor{violet!15}
      \quad +\textbf{SPARK} & 96.9 & 86.3 & 94.8 & 82.1 & 93.6 & 84.2 & \textbf{90.6}\,{\scriptsize\textcolor{teal!60!black}{(+2.6)}} \\
      \addlinespace[3pt]
      \cmidrule[0.35pt]{1-8}
      \rowcolor{blue!6}
      \QwenIcon\ 3-VL-8B & 99.3 & 88.6 & 97.9 & 88.6 & 96.6 & 87.2 & 93.7 \\
      \quad +MVP & 99.0 & 88.6 & 98.5 & 92.9 & 95.3 & 86.7 & 93.9\,{\scriptsize\textcolor{teal!60!black}{(+0.2)}} \\
      \rowcolor{violet!15}
      \quad +\textbf{SPARK} & 98.3 & 89.6 & 98.5 & 93.6 & 97.4 & 86.7 & \textbf{94.3}\,{\scriptsize\textcolor{teal!60!black}{(+0.6)}} \\
      \bottomrule
    \end{tabularx}
  \end{minipage}
  \par\vspace{8pt}
  \begin{minipage}[t]{0.49\textwidth}
    \centering
    \textbf{(c) OSWorld-G}\\[3pt]
    \begin{tabularx}{\linewidth}{@{}l >{\hsize=0.88\hsize\linewidth=\hsize\columncolor{gray!5}\centering\arraybackslash}X >{\hsize=0.88\hsize\linewidth=\hsize\centering\arraybackslash}X >{\hsize=0.88\hsize\linewidth=\hsize\columncolor{gray!5}\centering\arraybackslash}X >{\hsize=0.88\hsize\linewidth=\hsize\centering\arraybackslash}X >{\hsize=0.88\hsize\linewidth=\hsize\columncolor{gray!5}\centering\arraybackslash}X >{\hsize=1.6\hsize\linewidth=\hsize\centering\arraybackslash}X@{}}
      \toprule
      \textbf{Model / Method} & Text & Elem. & Layout & Fine & Ref. & \textbf{Avg.}$\uparrow$ \\
      \midrule
      \rowcolor{blue!6}
      \UITARSIcon 1.5-7B & 77.8 & 68.2 & 68.8 & 57.0 & $\text{N/A}$ & 65.6 \\
      \quad +MVP & 75.1 & 70.0 & 73.5 & 52.3 & $\text{N/A}$ & 66.0\,{\scriptsize\textcolor{teal!60!black}{(+0.4)}} \\
      \rowcolor{violet!15}
      \quad +\textbf{SPARK} & 79.3 & 74.5 & 76.7 & 59.1 & $\text{N/A}$ & \textbf{70.2}\,{\scriptsize\textcolor{teal!60!black}{(+4.6)}} \\
      \addlinespace[3pt]
      \cmidrule[0.35pt]{1-7}
      \rowcolor{blue!6}
      \QwenIcon 2.5-VL-7B & 55.9 & 46.4 & 53.4 & 28.9 & $\text{N/A}$ & 45.6 \\
      \quad +MVP & 67.4 & 61.5 & 66.8 & 40.9 & $\text{N/A}$ & 58.2\,{\scriptsize\textcolor{teal!60!black}{(+12.6)}} \\
      \rowcolor{violet!15}
      \quad +\textbf{SPARK} & 69.7 & 62.4 & 66.8 & 40.9 & $\text{N/A}$ & \textbf{59.0}\,{\scriptsize\textcolor{teal!60!black}{(+13.4)}} \\
      \addlinespace[3pt]
      \cmidrule[0.35pt]{1-7}
      \rowcolor{blue!6}
      \QwenIcon 3-VL-8B & 80.1 & 71.5 & 74.3 & 59.1 & $\text{N/A}$ & 68.9 \\
      \quad +MVP & 80.1 & 78.2 & 81.4 & 58.4 & $\text{N/A}$ & 72.6\,{\scriptsize\textcolor{teal!60!black}{(+3.7)}} \\
      \rowcolor{violet!15}
      \quad +\textbf{SPARK} & 83.1 & 79.4 & 82.6 & 62.4 & $\text{N/A}$ & \textbf{74.6}\,{\scriptsize\textcolor{teal!60!black}{(+5.7)}} \\
      \bottomrule
    \end{tabularx}
  \end{minipage}\hfill%
  \begin{minipage}[t]{0.49\textwidth}
    \centering
    \textbf{(d) UI-Vision}\\[3pt]
    \begin{tabularx}{\linewidth}{@{}l >{\hsize=0.92\hsize\linewidth=\hsize\columncolor{gray!5}\centering\arraybackslash}X >{\hsize=0.92\hsize\linewidth=\hsize\centering\arraybackslash}X >{\hsize=0.92\hsize\linewidth=\hsize\columncolor{gray!5}\centering\arraybackslash}X >{\hsize=1.24\hsize\linewidth=\hsize\centering\arraybackslash}X@{}}
      \toprule
      \textbf{Model / Method} & Basic & Func. & Spatial & \textbf{Avg.}$\uparrow$ \\
      \midrule
      \rowcolor{blue!6}
      \UITARSIcon 1.5-7B & 29.3 & 27.7 & 9.6 & 22.2 \\
      \quad +MVP & 31.7 & 30.9 & 13.4 & 25.3\,{\scriptsize\textcolor{teal!60!black}{(+3.1)}} \\
      \rowcolor{violet!15}
      \quad +\textbf{SPARK} & 34.7 & 33.9 & 15.2 & \textbf{28.0}\,{\scriptsize\textcolor{teal!60!black}{(+5.8)}} \\
      \addlinespace[3pt]
      \cmidrule[0.35pt]{1-5}
      \rowcolor{blue!6}
      \QwenIcon 2.5-VL-7B & 16.2 & 17.9 & 7.3 & 13.8 \\
      \quad +MVP & 22.1 & 26.5 & 10.3 & 19.6\,{\scriptsize\textcolor{teal!60!black}{(+5.8)}} \\
      \rowcolor{violet!15}
      \quad +\textbf{SPARK} & 24.9 & 28.1 & 10.1 & \textbf{21.0}\,{\scriptsize\textcolor{teal!60!black}{(+7.2)}} \\
      \addlinespace[3pt]
      \cmidrule[0.35pt]{1-5}
      \rowcolor{blue!6}
      \QwenIcon 3-VL-8B & 32.8 & 34.7 & 14.3 & 27.3 \\
      \quad +MVP & 38.5 & 39.7 & 19.6 & 32.6\,{\scriptsize\textcolor{teal!60!black}{(+5.3)}} \\
      \rowcolor{violet!15}
      \quad +\textbf{SPARK} & 39.4 & 40.8 & 19.3 & \textbf{33.2}\,{\scriptsize\textcolor{teal!60!black}{(+5.9)}} \\
      \bottomrule
    \end{tabularx}
  \end{minipage}
\end{table*}

\paragraph{Grounding complex professional interfaces.}
Table~\ref{tab:sspro} demonstrates the effectiveness of SPARK on ScreenSpot-Pro.
It achieves grounding accuracies of 68.7\%, 52.5\%, and 58.2\% with Qwen3-VL, Qwen2.5-VL, and UI-TARS, respectively, improving over single-view inference by 13.9, 25.2, and 16.1 pp.
The average improvement across the three backbones is 18.4 pp.
These gains span both general-purpose VLMs and a GUI-specialized model, showing that semantic inspection can improve grounding across different levels of backbone specialization.

The category-level results highlight the benefits for visually demanding interface elements.
With Qwen3-VL, SPARK raises icon-grounding accuracy from 14.1\% to 40.6\% in CAD, from 35.8\% to 67.9\% in Office, and from 27.0\% to 56.2\% in OS.
The improvements across these categories support the value of observations that expose distinguishing details together with their relevant context.

\paragraph{Benefits across benchmarks and backbones.}
Table~\ref{tab:grounding-four-benchmarks} extends the evaluation to ScreenSpot-v1, ScreenSpot-v2,
OSWorld-G, and UI-Vision.
SPARK improves overall accuracy over single-view inference in all benchmark--backbone settings.
With Qwen3-VL, it reaches 91.1\% on ScreenSpot-v1 and 94.3\% on ScreenSpot-v2, demonstrating further gains on top of strong initial grounding performance.

The benefits also extend to OSWorld-G and UI-Vision.
On OSWorld-G, SPARK reaches 74.6\%, 59.0\%, and 70.2\% with Qwen3-VL, Qwen2.5-VL, and UI-TARS, respectively.
On UI-Vision, it improves the corresponding single-view accuracies from 27.3\% to 33.2\%, from 13.8\% to 21.0\%, and from 22.2\% to 28.0\%.
Together, these results demonstrate the applicability of probe-guided observation across professional, desktop, mobile, and web interfaces.

\begin{figure*}[!t]
  \centering

  % 表格
  \sbox{\probeablationbox}{%
    \scriptsize
    \setlength{\tabcolsep}{4pt}
    \resizebox{0.48\textwidth}{!}{%
      \begin{tabular}{@{}l
        >{\columncolor{gray!5}}r
        r
        >{\columncolor{gray!5}}r@{}}
        \toprule
        Configuration
          & \UITARSIcon\ 1.5-7B
          & \QwenIcon\ 2.5-VL-7B
          & \QwenIcon\ 3-VL-8B \\
        \midrule
        \rowcolor{blue!6}
        Single-view baseline
          & 42.1 & 27.3 & 54.8 \\
        \midrule
        \quad w/o Probe Specialization
          & 57.5 & 51.7 & 68.3 \\
        \quad w/o Probe Hierarchy
          & 57.4 & 51.8 & 68.3 \\
        \quad w/o Coverage Discount
          & 58.0 & 50.7 & 67.6 \\
        \quad w/o Attention-Based Selection
          & 54.5 & 46.6 & 68.1 \\
        \midrule
        \rowcolor{violet!15}
        \textbf{SPARK}
          & \textbf{58.2} & \textbf{52.5} & \textbf{68.7} \\
        \bottomrule
      \end{tabular}%
    }%
  }

  % 图片：占满左侧宽度
  \sbox{\qwenfigurebox}{%
    \includegraphics[width=0.48\textwidth]{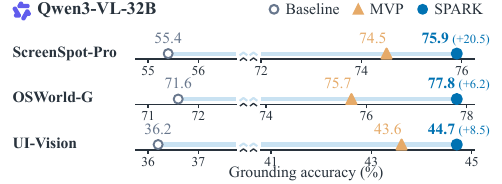}%
  }

  % 左侧整体：图片 + 下方 caption
  \sbox{\qwenpanelbox}{%
    \begin{minipage}{0.48\textwidth}
      \centering
      \usebox{\qwenfigurebox}
      \par
      \captionof{figure}{Grounding accuracy (\%) with
        Qwen3-VL-32B.}
      \label{fig:qwen3-32b}
    \end{minipage}%
  }

  % 右侧整体：上方 caption + 表格
  \sbox{\ablationpanelbox}{%
    \begin{minipage}{0.48\textwidth}
      \centering
      \captionof{table}{Ablation of SPARK on
        ScreenSpot-Pro. Accuracy (\%).}
      \label{tab:probe-ablation}
      \usebox{\probeablationbox}
    \end{minipage}%
  }

  % 统一两侧整体高度
  \setlength{\pairedpanelheight}{%
    \dimexpr\ht\qwenpanelbox+\dp\qwenpanelbox\relax
  }
  \ifdim
    \dimexpr\ht\ablationpanelbox+\dp\ablationpanelbox\relax
    >\pairedpanelheight
    \setlength{\pairedpanelheight}{%
      \dimexpr\ht\ablationpanelbox+\dp\ablationpanelbox\relax
    }
  \fi

  % 左侧表格：caption 在上
  \begin{minipage}[t][\pairedpanelheight][t]{0.48\textwidth}
    \vspace{0pt}
    \noindent\usebox{\ablationpanelbox}
  \end{minipage}%
  \hfill
  % 右侧图片：caption 在下
  \begin{minipage}[t][\pairedpanelheight][t]{0.48\textwidth}
    \vspace{0pt}
    \noindent\usebox{\qwenpanelbox}
  \end{minipage}
\vspace{-0.1in}
\end{figure*}

\vspace{-0.05in}
\paragraph{Scaling to a larger backbone.}
Figure~\ref{fig:qwen3-32b} shows that SPARK's benefits extend to Qwen3-VL-32B.
It achieves 75.9\% on ScreenSpot-Pro, 77.8\% on OSWorld-G, and 44.7\% on UI-Vision, improving over the single-view baseline.
SPARK also consistently outperforms MVP.
These results show that semantic probe-guided observation remains effective at a larger model scale, complementing increased backbone capacity with task-relevant inspection.
Full results are provided in Appendix Table 6.

\vspace{-0.05in}
\paragraph{Comparison with test-time grounding methods.}
SPARK achieves the highest ScreenSpot-Pro accuracy among the compared test-time methods for each backbone.
In particular, it surpasses zoom-based approaches:
with Qwen2.5-VL, its 52.5\% accuracy exceeds RegionFocus and DRS-GUI by 20.4 and 11.6 pp, respectively.
With UI-TARS, SPARK reaches 58.2\%, compared with 51.9\% for BAMI.
Against the SOTA multi-view method MVP,
SPARK improves overall accuracy in all evaluated settings.
The gains on ScreenSpot-Pro are 3.5 pp for Qwen3-VL, 3.4 pp for Qwen2.5-VL, and 2.3 pp for UI-TARS.
These gains demonstrate the effectiveness of semantic probe-guided observation over existing test-time grounding strategies.

\subsection{Ablation and Analysis}

\paragraph{Contributions of probe-guided view construction.}
Table~\ref{tab:probe-ablation} examines how probe design
and spatial guidance contribute to view construction.
Without probe specialization, verification is replaced
by an additional exploration view under the same
four-subview budget.
Restoring verification improves accuracy across all
three backbones.
Hierarchical descriptions also consistently outperform
unified descriptions, supporting the separation of
contextual and element-level evidence.

To examine how semantic guidance is spatially grounded,
we replace attention-based selection with directly
predicted crop coordinates, retaining the same maximum
local-view budget and final aggregation.
Attention-based selection improves accuracy across
all three backbones, supporting spatial attention as
an effective interface between semantic descriptions
and visual observations.
Coverage discounting further improves performance,
showing the value of accounting for observation history
when selecting complementary views.

\begin{figure*}[!t]
    \centering
    \includegraphics[width=1\linewidth]{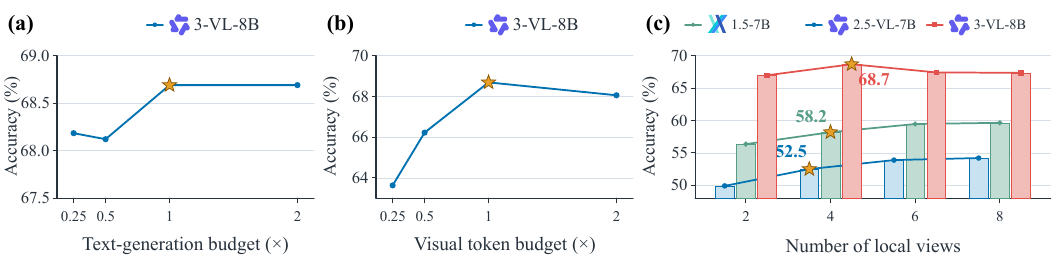}
    \vspace{-0.15in}
    \caption{Budget scaling on ScreenSpot-Pro.
    (a) Probe text-generation budget.
    (b) Visual token budget.
    (c) Number of local views.}
    \label{fig:abl-scaling}
\vspace{-0.05in}
\end{figure*}

\vspace{-0.1in}
\paragraph{Contributions of adaptive observation scope.}
Table~\ref{tab:scale-ablation} compares adaptive focusing with fixed crop scales under matched visual token budgets.
Adaptive selection achieves 68.7\% accuracy, outperforming all fixed-scale settings.
Its average crop area lies between the fixed $1.0\times$ and $1.4\times$ settings, indicating that the gain comes from selecting an appropriate contextual scope for each observation rather than uniformly enlarging the crops. 
These results support using contextual evidence to adapt the extent of visual inspection.

\begin{table}[!t]
  \centering
  \caption{Effect of adaptive scope on SSPro with Qwen3-VL-8B.}
  \vspace{-0.05in}
  \label{tab:scale-ablation}
  \scriptsize
  \setlength{\tabcolsep}{4pt}
  \resizebox{\columnwidth}{!}{%
    \begin{tabular}{@{}l
      >{\columncolor{gray!5}}r
      r
      >{\columncolor{gray!5}}r
      >{\columncolor{violet!15}}r@{}}
      \toprule
      Scale setting
        & Fixed $0.7\times$
        & Fixed $1.0\times$
        & Fixed $1.4\times$
        & \textbf{Adaptive} \\
      \midrule
      Acc.$\uparrow$ (\%)
        & 65.0 & 67.0 & 67.7 & \textbf{68.7} \\
      Crop area (\%)
        & 12.0 & 19.5 & 33.5 & 23.7 \\
      \bottomrule
    \end{tabular}%
  }
\end{table}

\vspace{-0.05in}
\paragraph{Contributions of uncertainty-weighted aggregation.}
Table~\ref{tab:selection-ablation} isolates the effect of
aggregation on ScreenSpot-Pro using identical SPARK-generated
observations and predictions within each backbone.
Uncertainty-weighted voting achieves the highest accuracy
among the evaluated rules, reaching 58.2\%, 52.5\%, and 68.7\%
with UI-TARS, Qwen2.5-VL, and Qwen3-VL, respectively.
It outperforms majority voting used in MVP~\cite{mvp}
by 1.9, 1.5, and 1.7 pp, and minimum-entropy
selection by 1.6, 1.5, and 2.9 pp.

These results support combining predictive uncertainty with
spatial agreement.
Minimum-entropy selection retains only one prediction,
while majority voting weights all predictions equally.
Our rule instead accumulates entropy-weighted support within
spatial clusters and returns the medoid of the highest-scoring
cluster.
The consistent gains across general-purpose and GUI-specialized
backbones demonstrate the benefit of uncertainty-weighted
aggregation without additional observations.

\vspace{-0.05in}
\paragraph{Scaling text and visual token budgets.}
Figures~\ref{fig:abl-scaling}(a,b) examine semantic
description and local visual token budgets with
Qwen3-VL on ScreenSpot-Pro.
Accuracy remains between 68.1\% and 68.7\% as probe
generation caps vary from $0.25\times$ to $2\times$,
showing that concise descriptions provide effective
inspection guidance.
In contrast, increasing the visual token budget
from $0.25\times$ to $1\times$ raises accuracy from
63.6\% to 68.7\%.
The default setting, with up to 3,680 visual tokens
per local view, achieves the highest accuracy among
the evaluated budgets.
These results support pairing compact semantic guidance
with sufficient visual detail to resolve the identified evidence.

\vspace{-0.05in}
\paragraph{Scaling the number of local observations.}
Figure~\ref{fig:abl-scaling}(c) evaluates observation
budgets from two to eight local views.
Increasing the budget from two to four improves accuracy
across all three backbones, reaching 68.7\% for Qwen3-VL,
52.5\% for Qwen2.5-VL, and 58.2\% for UI-TARS.
Four views achieve the best result for Qwen3-VL,
while additional exploration raises Qwen2.5-VL and
UI-TARS to 54.2\% and 59.6\% at eight views.
The smaller gains from six to eight views indicate
diminishing returns, supporting four views as an
effective default while allowing further improvements
through a larger observation budget.

\begin{table}[!t]
  \centering
  \caption{Effect of uncertainty-weighted voting on SSPro.}
  \vspace{-0.05in}
  \label{tab:selection-ablation}
  \scriptsize
  \setlength{\tabcolsep}{4pt}
  \resizebox{\columnwidth}{!}{%
    \begin{tabular}{@{}l
      >{\columncolor{gray!5}}r
      r
      >{\columncolor{gray!5}}r@{}}
      \toprule
      Selection rule
        & \UITARSIcon\ 1.5-7B
        & \QwenIcon\ 2.5-VL-7B
        & \QwenIcon\ 3-VL-8B \\
      \midrule
      Random selection
        & 41.6 & 36.4 & 54.9 \\
      Last view
        & 43.3 & 38.5 & 58.1 \\
      Minimum-entropy selection
        & 56.6 & 51.0 & 65.8 \\
      Majority voting~\cite{mvp}
        & 56.3 & 51.0 & 67.0 \\
      \rowcolor{violet!15}
      \textbf{Uncertainty-weighted voting}
        & \textbf{58.2} & \textbf{52.5} & \textbf{68.7} \\
      \bottomrule
    \end{tabular}%
  }
\end{table}

\section{Conclusion}
\label{sec:conclusion}

We presented \textbf{SPARK}, a training-free framework that turns
a frozen VLM's semantic understanding into guidance for
visual inspection.
Hierarchical probes support exploration and
hypothesis-conditioned verification, translating
descriptions of relevant evidence into spatial attention
that guides view placement and contextual scope.
History-aware selection encourages complementary local
views, while uncertainty-weighted voting integrates
their predictions.
Experiments across GUI grounding benchmarks and backbones
demonstrate consistent improvements over single-view
inference and strong performance against existing
test-time methods.
SPARK shows that \textit{knowing what to inspect can help a VLM
determine where to act}, connecting semantic understanding
with precise GUI grounding.

% \input{sec/1_intro}
% \input{sec/2_formatting}
% \input{sec/3_finalcopy}
% \clearpage

\section*{LLM Usage Statement}

The authors used generative AI tools, specifically GPT and Codex powered by GPT models, to assist with writing, LaTeX and figure formatting, and code editing. All claims, citations, code, figures, and numerical results were independently reviewed and verified by the authors, who take full responsibility for the final submission.

{
    \small
    \bibliographystyle{ieeenat_fullname}
    \bibliography{main}

@String(CVPR= {IEEE Conf. Comput. Vis. Pattern Recog.})

@String(AAAI = {AAAI})

@String(CVPR  = {CVPR})

@inproceedings{nguyen2025gui,
  title={Gui agents: A survey},
  author={Nguyen, Dang and Chen, Jian and Wang, Yu and Wu, Gang and Park, Namyong and Hu, Zhengmian and Lyu, Hanjia and Wu, Junda and Aponte, Ryan and Xia, Yu and others},
  booktitle={Findings of the Association for Computational Linguistics: ACL 2025},
  pages={22522--22538},
  year={2025}
}

@inproceedings{seeclick,
  title     = {{SeeClick}: Harnessing {GUI} Grounding for Advanced Visual {GUI} Agents},
  author    = {Cheng, Kanzhi and Sun, Qiushi and Chu, Yougang and
               Xu, Fangzhi and Li, Yantao and Zhang, Jianbing and
               Wu, Zhiyong},
  booktitle = {Proceedings of the 62nd Annual Meeting of the Association
               for Computational Linguistics (Volume 1: Long Papers)},
  year      = {2024},
  pages     = {9313--9332},
}

@inproceedings{uground,
  title     = {Navigating the Digital World as Humans Do:
               Universal Visual Grounding for {GUI} Agents},
  author    = {Gou, Boyu and Wang, Ruohan and Zheng, Boyuan and
               Xie, Yanan and Chang, Cheng and Shu, Yiheng and
               Sun, Huan and Su, Yu},
  booktitle = {The Thirteenth International Conference on Learning Representations},
  year      = {2025},
}

@article{uitars,
  title   = {{UI-TARS}: Pioneering Automated {GUI} Interaction with Native Agents},
  author  = {Qin, Yujia and Ye, Yining and Fang, Junjie and
             Wang, Haoming and Liang, Shihao and Tian, Shizuo and
             Zhang, Junda and Li, Jiahao and Li, Yunxin and
             Huang, Shijue and Zhong, Wanjun and Li, Kuanye and
             Yang, Jiale and Miao, Yu and Lin, Woyu and
             Liu, Longxiang and Jiang, Xu and Ma, Qianli and
             Li, Jingyu and Xiao, Xiaojun and Cai, Kai and
             Li, Chuang and Zheng, Yaowei and Jin, Chaolin and
             Li, Chen and Zhou, Xiao and Wang, Minchao and
             Chen, Haoli and Li, Zhaojian and Yang, Haihua and
             Liu, Haifeng and Lin, Feng and Peng, Tao and
             Liu, Xin and Shi, Guang},
  journal = {arXiv preprint arXiv:2501.12326},
  year    = {2025},
  url     = {https://arxiv.org/abs/2501.12326}
}

@inproceedings{tag,
  title     = {Attention-Driven {GUI} Grounding: Leveraging Pretrained
               Multimodal Large Language Models Without Fine-Tuning},
  author    = {Xu, Hai-Ming and Chen, Qi and Wang, Lei and Liu, Lingqiao},
  booktitle = {Proceedings of the AAAI Conference on Artificial Intelligence},
  volume    = {39},
  number    = {8},
  pages     = {8851--8859},
  year      = {2025},
}

@inproceedings{trifuse,
  title     = {{Trifuse}: Enhancing Attention-Based {GUI} Grounding
               via Multimodal Fusion},
  author    = {Ma, Longhui and Zhao, Di and Wang, Siwei and
               Lv, Zhao and Wang, Miao},
  booktitle = {Proceedings of the 43rd International Conference on Machine Learning},
  series    = {Proceedings of Machine Learning Research},
  volume    = {306},
  pages     = {85344--85363},
  year      = {2026},
}

@inproceedings{regionfocus,
  title     = {Visual Test-time Scaling for {GUI} Agent Grounding},
  author    = {Luo, Tiange and Logeswaran, Lajanugen and
               Johnson, Justin and Lee, Honglak},
  booktitle = {Proceedings of the IEEE/CVF International Conference
               on Computer Vision},
  pages     = {19989--19998},
  year      = {2025},
}

@inproceedings{dimogui,
  title     = {{DiMo-GUI}: Advancing Test-time Scaling in {GUI}
               Grounding via Modality-Aware Visual Reasoning},
  author    = {Wu, Hang and Chen, Hongkai and Cai, Yujun and
               Liu, Chang and Ye, Qingwen and Yang, Ming-Hsuan and
               Wang, Yiwei},
  booktitle = {Proceedings of the 2025 Conference on Empirical
               Methods in Natural Language Processing},
  pages     = {26246--26256},
  year      = {2025},
}

@inproceedings{mvp,
  title={{MVP}: Multiple view prediction improves gui grounding},
  author={Zhang, Yunzhu and Pan, Zeyu and Zeng, Zhengwen and Shen, Shuheng and Meng, Changhua and Zhu, Linchao},
  booktitle={Proceedings of the IEEE/CVF Conference on Computer Vision and Pattern Recognition},
  pages={27482--27492},
  year={2026}
}

@inproceedings{drsgui,
  title     = {{DRS-GUI}: Dynamic Region Search for Training-Free {GUI} Grounding},
  author    = {Liu, Yichao and Shen, Huawen and Yu, Liu and
               Liu, Shiyu and Chen, Zeyu and Zhou, Yu},
  booktitle = {Proceedings of the IEEE/CVF Conference on
               Computer Vision and Pattern Recognition},
  pages     = {34606--34616},
  year      = {2026},
}

@inproceedings{vstar,
  title     = {{V*}: Guided Visual Search as a Core Mechanism
               in Multimodal {LLMs}},
  author    = {Wu, Penghao and Xie, Saining},
  booktitle = {Proceedings of the IEEE/CVF Conference on
               Computer Vision and Pattern Recognition},
  pages     = {13084--13094},
  year      = {2024},
}

@inproceedings{visualcot,
  title     = {{Visual CoT}: Advancing Multi-Modal Language Models
               with a Comprehensive Dataset and Benchmark
               for Chain-of-Thought Reasoning},
  author    = {Shao, Hao and Qian, Shengju and Xiao, Han and
               Song, Guanglu and Zong, Zhuofan and Wang, Letian and
               Liu, Yu and Li, Hongsheng},
  booktitle = {Advances in Neural Information Processing Systems},
  volume    = {37},
  pages     = {8612--8642},
  year      = {2024},
}

@inproceedings{visrl,
  title     = {{VisRL}: Intention-Driven Visual Perception
               via Reinforced Reasoning},
  author    = {Chen, Zhangquan and Luo, Xufang and Li, Dongsheng},
  booktitle = {Proceedings of the IEEE/CVF International Conference
               on Computer Vision},
  pages     = {2545--2555},
  year      = {2025},
}

@InProceedings{Zhang2026bami,
    author    = {Zhang, Borui and Zhang, Bo and Wang, Bo and Zheng, Wenzhao and Cheng, Yuhao and Tang, Liang and Yan, Yiqiang and Zhou, Jie and Lu, Jiwen},
    title     = {{BAMI}: Training-Free Bias Mitigation in GUI Grounding},
    booktitle = {Proceedings of the IEEE/CVF Conference on Computer Vision and Pattern Recognition (CVPR)},
    year      = {2026},
    pages     = {34596-34605}
}

@inproceedings{li2025screenspot,
  title={Screenspot-pro: Gui grounding for professional high-resolution computer use},
  author={Li, Kaixin and Meng, Ziyang and Lin, Hongzhan and Luo, Ziyang and Tian, Yuchen and Ma, Jing and Huang, Zhiyong and Chua, Tat-Seng},
  booktitle={Proceedings of the 33rd ACM International Conference on Multimedia},
  pages={8778--8786},
  year={2025}
}

@inproceedings{osatlas,
  title     = {{OS-ATLAS}: Foundation Action Model
               for Generalist {GUI} Agents},
  author    = {Wu, Zhiyong and Wu, Zhenyu and Xu, Fangzhi and
               Wang, Yian and Sun, Qiushi and Jia, Chengyou and
               Cheng, Kanzhi and Ding, Zichen and Chen, Liheng and
               Liang, Paul Pu and Qiao, Yu},
  booktitle = {The Thirteenth International Conference
               on Learning Representations},
  year      = {2025},
}

@misc{uitars15,
  title        = {{UI-TARS-1.5-7B}},
  author       = {{ByteDance Seed}},
  howpublished = {Hugging Face model card},
  year         = {2025},
  url          = {https://huggingface.co/ByteDance-Seed/UI-TARS-1.5-7B},
}

@inproceedings{uivision,
  title     = {{UI-Vision}: A Desktop-centric {GUI} Benchmark
               for Visual Perception and Interaction},
  author    = {Nayak, Shravan and Jian, Xiangru and
               Lin, Kevin Qinghong and Rodriguez, Juan A. and
               Kalsi, Montek and Chapados, Nicolas and
               {\"O}zsu, M. Tamer and Agrawal, Aishwarya and
               Vazquez, David and Pal, Christopher and
               Taslakian, Perouz and Gella, Spandana and
               Rajeswar, Sai},
  booktitle = {Proceedings of the 42nd International Conference
               on Machine Learning},
  volume    = {267},
  pages     = {45817--45851},
  year      = {2025},
}

@inproceedings{osworldg,
  title     = {Scaling Computer-Use Grounding via
               User Interface Decomposition and Synthesis},
  author    = {Xie, Tianbao and Deng, Jiaqi and Li, Xiaochuan and
               Yang, Junlin and Wu, Haoyuan and Chen, Jixuan and
               Hu, Wenjing and Wang, Xinyuan and Xu, Yuhui and
               Wang, Zekun and Xu, Yiheng and Wang, Junli and
               Sahoo, Doyen and Yu, Tao and Xiong, Caiming},
  booktitle = {Advances in Neural Information Processing Systems},
  year      = {2025},
  url       = {https://arxiv.org/abs/2505.13227}
}

@article{qwen3vl,
  title   = {{Qwen3-VL} Technical Report},
  author  = {Bai, Shuai and Cai, Yuxuan and Chen, Ruizhe and
             Chen, Keqin and Chen, Xionghui and Cheng, Zesen and
             Deng, Lianghao and Ding, Wei and Gao, Chang and
             Ge, Chunjiang and Ge, Wenbin and Guo, Zhifang and
             Huang, Qidong and Huang, Jie and Huang, Fei and
             Hui, Binyuan and Jiang, Shutong and Li, Zhaohai and
             Li, Mingsheng and Li, Mei and Li, Kaixin and
             Lin, Zicheng and Lin, Junyang and Liu, Xuejing and
             Liu, Jiawei and Liu, Chenglong and Liu, Yang and
             Liu, Dayiheng and Liu, Shixuan and Lu, Dunjie and
             Luo, Ruilin and Lv, Chenxu and Men, Rui and
             Meng, Lingchen and Ren, Xuancheng and Ren, Xingzhang and
             Song, Sibo and Sun, Yuchong and Tang, Jun and
             Tu, Jianhong and Wan, Jianqiang and Wang, Peng and
             Wang, Pengfei and Wang, Qiuyue and Wang, Yuxuan and
             Xie, Tianbao and Xu, Yiheng and Xu, Haiyang and
             Xu, Jin and Yang, Zhibo and Yang, Mingkun and
             Yang, Jianxin and Yang, An and Yu, Bowen and
             Zhang, Fei and Zhang, Hang and Zhang, Xi and
             Zheng, Bo and Zhong, Humen and Zhou, Jingren and
             Zhou, Fan and Zhou, Jing and Zhu, Yuanzhi and Zhu, Ke},
  journal = {arXiv preprint arXiv:2511.21631},
  year    = {2025},
  url     = {https://arxiv.org/abs/2511.21631}
}

@article{qwen25vl,
  title   = {{Qwen2.5-VL} Technical Report},
  author  = {Bai, Shuai and Chen, Keqin and Liu, Xuejing and
             Wang, Jialin and Ge, Wenbin and Song, Sibo and
             Dang, Kai and Wang, Peng and Wang, Shijie and
             Tang, Jun and Zhong, Humen and Zhu, Yuanzhi and
             Yang, Mingkun and Li, Zhaohai and Wan, Jianqiang and
             Wang, Pengfei and Ding, Wei and Fu, Zheren and
             Xu, Yiheng and Ye, Jiabo and Zhang, Xi and
             Xie, Tianbao and Cheng, Zesen and Zhang, Hang and
             Yang, Zhibo and Xu, Haiyang and Lin, Junyang},
  journal = {arXiv preprint arXiv:2502.13923},
  year    = {2025},
  url     = {https://arxiv.org/abs/2502.13923}
}

@article{hurst2024gpt,
  title={Gpt-4o system card},
  author={Hurst, Aaron and Lerer, Adam and Goucher, Adam P and Perelman, Adam and Ramesh, Aditya and Clark, Aidan and Ostrow, AJ and Welihinda, Akila and Hayes, Alan and Radford, Alec and others},
  journal={arXiv preprint arXiv:2410.21276},
  year={2024}
}

@article{hu2024dawn,
  title={The dawn of gui agent: A preliminary case study with claude 3.5 computer use},
  author={Hu, Siyuan and Ouyang, Mingyu and Gao, Difei and Shou, Mike Zheng},
  journal={arXiv preprint arXiv:2411.10323},
  year={2024}
}

@inproceedings{yuan2025enhancing,
 author = {Yuan, Xinbin and Zhang, Jian and Li, Kaixin and Cai, Zhuoxuan and Yao, Lujian and Chen, Jie and Wang, Enguang and Hou, Qibin and Chen, Jinwei and Jiang, Peng-Tao and Li, Bo},
 booktitle = {Advances in Neural Information Processing Systems},
 pages = {127658--127679},
 title = {SE-GUI: Enhancing Visual Grounding for GUI Agents via Self-Evolutionary Reinforcement Learning},
 year = {2025}
}

@inproceedings{tang2026gui,
  title={{GUI-G$^2$}: Gaussian Reward Modeling for {GUI} Grounding},
  author={Tang, Fei and Gu, Zhangxuan and Lu, Zhengxi and Liu, Xuyang and Shen, Shuheng and Meng, Changhua and Wang, Wen and Zhang, Wenqi and Shen, Yongliang and Lu, Weiming and others},
  booktitle={Proceedings of the AAAI Conference on Artificial Intelligence},
  volume={40},
  number={39},
  pages={33214--33222},
  year={2026}
}

@article{yao2026autofocus,
  title   = {{AutoFocus}: Uncertainty-Aware Active Visual Search for {GUI} Grounding},
  author  = {Yao, Ruilin and Xiong, Shengwu and Zou, Tianyu and
             Xiong, Shili and Rong, Yi},
  journal = {arXiv preprint arXiv:2605.02630},
  year    = {2026},
  url     = {https://arxiv.org/abs/2605.02630}
}

@article{chen2026mhvf,
  title   = {{MHVF}: Training-Free {GUI} Grounding via
             Verification-Guided Multi-Hypothesis Search},
  author  = {Chen, Minghua and Liao, Xingming and Zhong, Weiqiang and
             Wang, Zhuowei},
  journal = {arXiv preprint arXiv:2610.17792},
  year    = {2026},
  url     = {https://arxiv.org/abs/2610.17792}
}

@article{tang2025survey,
  title={A survey on (m) llm-based gui agents},
  author={Tang, Fei and Xu, Haolei and Zhang, Hang and Chen, Siqi and Wu, Xingyu and Shen, Yongliang and Zhang, Wenqi and Hou, Guiyang and Tan, Zeqi and Yan, Yuchen and others},
  journal={arXiv preprint arXiv:2504.13865},
  year={2025}
}

@article{tang2026uizoomer,
  title   = {{UI-Zoomer}: Uncertainty-Driven Adaptive Zoom-In
             for {GUI} Grounding},
  author  = {Tang, Fei and Chen, Bofan and Lu, Zhengxi
             and Chen, Tongbo and Nong, Songqin and Jiang, Tao
             and Xu, Wenhao and Lu, Weiming and Xiao, Jun
             and Zhuang, Yueting and Shen, Yongliang},
  journal = {arXiv preprint arXiv:2604.14113},
  year    = {2026}
}

@article{wang2026adaground,
  title   = {{AdaGround}: Token-Normalized Adaptive Zooming
             with Verifier-in-the-Loop Refinement
             for {GUI} Grounding},
  author  = {Wang, Bingchao and Jing, Yongcheng and Wang, Chengyu
             and Huang, Jun and Yang, Jie and Liu, Wei
             and Tao, Dacheng},
  journal = {arXiv preprint arXiv:2610.36874},
  year    = {2026}
}

@inproceedings{Xie2024OSWorld,
  title = {{OSWorld}: Benchmarking Multimodal Agents for Open-Ended Tasks in Real Computer Environments},
  author = {Xie, Tianbao and Zhang, Danyang and Chen, Jixuan and Li, Xiaochuan and Zhao, Siheng and Cao, Ruisheng and Hua, Toh Jing and Cheng, Zhoujun and Shin, Dongchan and Lei, Fangyu and Liu, Yitao and Xu, Yiheng and Zhou, Shuyan and Savarese, Silvio and Xiong, Caiming and Zhong, Victor and Yu, Tao},
  booktitle = {Advances in Neural Information Processing Systems},
  volume = {37},
  year = {2024},
}

@inproceedings{Hong2024CogAgent,
  title = {{CogAgent}: A Visual Language Model for {GUI} Agents},
  author = {Hong, Wenyi and Wang, Weihan and Lv, Qingsong and Xu, Jiazheng and Yu, Wenmeng and Ji, Junhui and Wang, Yan and Wang, Zihan and Dong, Yuxiao and Ding, Ming and Tang, Jie},
  booktitle = {Proceedings of the IEEE/CVF Conference on Computer Vision and Pattern Recognition},
  pages = {14281--14290},
  year = {2024},
}

@inproceedings{Xu2025Aguvis,
  title = {{Aguvis}: Unified Pure Vision Agents for Autonomous {GUI} Interaction},
  author = {Xu, Yiheng and Wang, Zekun and Wang, Junli and Lu, Dunjie and Xie, Tianbao and Saha, Amrita and Sahoo, Doyen and Yu, Tao and Xiong, Caiming},
  booktitle = {Proceedings of the 42nd International Conference on Machine Learning},
  volume = {267},
  pages = {69772--69805},
  year = {2025},
}

@inproceedings{Wu2025GUIActor,
  title = {{GUI-Actor}: Coordinate-Free Visual Grounding for {GUI} Agents},
  author = {Wu, Qianhui and Cheng, Kanzhi and Yang, Rui and Zhang, Chaoyun and Yang, Jianwei and Jiang, Huiqiang and Mu, Jian and Peng, Baolin and Qiao, Bo and Tan, Reuben and Qin, Si and Liden, Lars and Lin, Qingwei and Zhang, Huan and Zhang, Tong and Zhang, Jianbing and Zhang, Dongmei and Gao, Jianfeng},
  booktitle = {Advances in Neural Information Processing Systems},
  volume = {38},
  year = {2025},
}

@inproceedings{Zhang2025MLLMsKnow,
  title = {{MLLMs} Know Where to Look: Training-free Perception of Small Visual Details with Multimodal {LLMs}},
  author = {Zhang, Jiarui and Khayatkhoei, Mahyar and Chhikara, Prateek and Ilievski, Filip},
  booktitle = {The Thirteenth International Conference on Learning Representations},
  year = {2025},
}

@inproceedings{Li2020Mapping,
  title = {Mapping Natural Language Instructions to Mobile {UI} Action Sequences},
  author = {Li, Yang and He, Jiacong and Zhou, Xin and Zhang, Yuan and Baldridge, Jason},
  booktitle = {Proceedings of the 58th Annual Meeting of the Association for Computational Linguistics},
  pages = {8198--8210},
  year = {2020},
  publisher = {Association for Computational Linguistics},
}

@inproceedings{Pasupat2018Mapping,
  title = {Mapping Natural Language Commands to Web Elements},
  author = {Pasupat, Panupong and Jiang, Tian-Shun and Liu, Evan and Guu, Kelvin and Liang, Percy},
  booktitle = {Proceedings of the 2018 Conference on Empirical Methods in Natural Language Processing},
  pages = {4970--4976},
  year = {2018},
  publisher = {Association for Computational Linguistics},
}

@inproceedings{Shen2025ZoomEye,
  title = {{ZoomEye}: Enhancing Multimodal {LLMs} with Human-Like Zooming Capabilities through Tree-Based Image Exploration},
  author = {Shen, Haozhan and Zhao, Kangjia and Zhao, Tiancheng and Xu, Ruochen and Zhang, Zilun and Zhu, Mingwei and Yin, Jianwei},
  booktitle = {Proceedings of the 2025 Conference on Empirical Methods in Natural Language Processing},
  pages = {6602--6618},
  year = {2025},
}

@article{zhang2024large,
  title={Large language model-brained gui agents: A survey},
  author={Zhang, Chaoyun and He, Shilin and Qian, Jiaxu and Li, Bowen and Li, Liqun and Qin, Si and Kang, Yu and Ma, Minghua and Liu, Guyue and Lin, Qingwei and others},
  journal={arXiv preprint arXiv:2411.18279},
  year={2024}
}

@inproceedings{yang2026gta1,
  title     = {{GTA1}: {GUI} Test-time Scaling Agent},
  author    = {Yang, Yan and
               Li, Dongxu and
               Dai, Yutong and
               Yang, Yuhao and
               Luo, Ziyang and
               Zhao, Zirui and
               Hu, Zhiyuan and
               Huang, Junzhe and
               Saha, Amrita and
               Chen, Zeyuan and
               Xu, Ran and
               Pan, Liyuan and
               Xiong, Caiming and
               Li, Junnan},
  booktitle = {The Fourteenth International Conference on Learning Representations},
  year      = {2026},

}
}

% % WARNING: do not forget to delete the supplementary pages from your submission 
\clearpage
\setcounter{page}{1}
\maketitlesupplementary
\appendix

\begin{table*}[ht]
  \centering
  \caption{Grounding accuracy (\%) with Qwen3-VL-32B. Best overall accuracy per benchmark is shown in \textbf{bold}. N/A indicates that the Refusal (Ref.) category is not evaluated under our protocol, consistent with MVP~\cite{mvp}}
  \label{tab:qwen3-32b-benchmarks}
  \small
  \setlength{\tabcolsep}{4pt}
  \renewcommand{\arraystretch}{1.12}

  % ---------------- (a) ScreenSpot-Pro ----------------
  \textbf{(a) ScreenSpot-Pro}\\[4pt]
  \resizebox{\textwidth}{!}{%
    \begin{tabular}{@{}l
      >{\columncolor{gray!5}}r >{\columncolor{gray!5}}r
      r r
      >{\columncolor{gray!5}}r >{\columncolor{gray!5}}r
      r r
      >{\columncolor{gray!5}}r >{\columncolor{gray!5}}r
      r r c@{}}
      \toprule
      \multirow{2}{*}{\textbf{Model / Method}}
        & \multicolumn{2}{c}{\cellcolor{gray!12}Dev.}
        & \multicolumn{2}{c}{Creative}
        & \multicolumn{2}{c}{\cellcolor{gray!12}CAD}
        & \multicolumn{2}{c}{Sci.}
        & \multicolumn{2}{c}{\cellcolor{gray!12}Office}
        & \multicolumn{2}{c}{OS}
        & \multirow{2}{*}{\shortstack{Avg.$\uparrow$\\[-1pt]{\scriptsize($\Delta$ pp)}}} \\
      \cmidrule(lr){2-3}\cmidrule(lr){4-5}\cmidrule(lr){6-7}
      \cmidrule(lr){8-9}\cmidrule(lr){10-11}\cmidrule(lr){12-13}
        & Text & Icon & Text & Icon & Text & Icon
        & Text & Icon & Text & Icon & Text & Icon & \\
      \midrule
      \QwenIcon\ Qwen3-VL-32B
        & 66.2 & 20.0 & 77.3 & 21.7 & 67.5 & 28.1
        & 84.7 & 33.6 & 80.8 & 49.1 & 59.8 & 20.2
        & 55.4 \\
      \quad +MVP
        & 90.3 & 51.0 & 84.8 & 47.6 & 85.8 & 46.9
        & 89.6 & 46.4 & 91.5 & 77.4 & 88.8 & 58.4
        & 74.5\,{\scriptsize\textcolor{teal!60!black}{(+19.1)}} \\
      \rowcolor{violet!15}
      \quad +SPARK
        & 92.9 & 51.0 & 86.9 & 51.0 & 84.3 & 46.9
        & 91.0 & 46.4 & 93.2 & 83.0 & 87.9 & 64.0
        & \textbf{75.9}\,{\scriptsize\textcolor{teal!60!black}{(+20.5)}} \\
      \bottomrule
    \end{tabular}%
  }

  \vspace{8pt}

  % ---------------- (b) OSWorld-G / (c) UI-Vision ----------------
  \resizebox{\textwidth}{!}{%
    \begin{tabular}{@{}l
      >{\columncolor{gray!5}}r
      r
      >{\columncolor{gray!5}}r
      r
      >{\columncolor{gray!5}}r
      r
      >{\columncolor{white}}p{10pt}
      l
      >{\columncolor{gray!5}}r
      r
      >{\columncolor{gray!5}}r
      r@{}}
      \multicolumn{7}{c}{\textbf{(b) OSWorld-G}}
        & & \multicolumn{5}{c}{\textbf{(c) UI-Vision}} \\[4pt]
      \cmidrule[\heavyrulewidth]{1-7}
      \cmidrule[\heavyrulewidth]{9-13}
      \textbf{Model / Method}
        & Text & Elem. & Layout & Fine & Ref.
        & Avg.$\uparrow$\,{\scriptsize($\Delta$ pp)}
        & & \textbf{Model / Method}
        & Basic & Func. & Spatial
        & Avg.$\uparrow$\,{\scriptsize($\Delta$ pp)} \\
      \cmidrule[\lightrulewidth]{1-7}
      \cmidrule[\lightrulewidth]{9-13}
      \QwenIcon\ Qwen3-VL-32B
        & 81.2 & 76.1 & 77.5 & 61.1 & N/A & 71.6
        & & \QwenIcon\ Qwen3-VL-32B
        & 43.6 & 42.9 & 22.3 & 36.2 \\
      \quad +MVP
        & 81.6 & 82.4 & 85.8 & 61.1 & N/A
        & 75.7\,{\scriptsize\textcolor{teal!60!black}{(+4.1)}}
        & & \quad +MVP
        & 50.2 & 49.7 & 30.9
        & 43.6\,{\scriptsize\textcolor{teal!60!black}{(+7.4)}} \\
      \rowcolor{violet!15}
      \quad +SPARK
        & 84.3 & 84.2 & 87.0 & 65.1 & N/A
        & \textbf{77.8}\,{\scriptsize\textcolor{teal!60!black}{(+6.2)}}
        & \cellcolor{white}
        & \quad +SPARK
        & 51.4 & 51.1 & 31.6
        & \textbf{44.7}\,{\scriptsize\textcolor{teal!60!black}{(+8.5)}} \\
      \cmidrule[\heavyrulewidth]{1-7}
      \cmidrule[\heavyrulewidth]{9-13}
    \end{tabular}%
  }
\end{table*}

\section{More Related Work}
\label{app:more-related-work}

Recent preprints investigate uncertainty-guided search
and visual verification for training-free GUI grounding.
UI-Zoomer~\cite{tang2026uizoomer} combines spatial
consensus and generation confidence to trigger zooming,
and adapts crop size to prediction variance.
AutoFocus~\cite{yao2026autofocus} converts sampled
coordinates and their axial perplexities into spatial
proposals, followed by visual candidate comparison.
MHVF~\cite{chen2026mhvf} explores multiple hypotheses
through zoomed re-localization and uses local
verification scores to guide further search and
final selection.
AdaGround~\cite{wang2026adaground} adapts zoom parameters
to estimated target size and uses a marked-click
verifier to guide acceptance and refinement.

SPARK derives observation guidance from explicit
descriptions of the evidence to inspect.
Attention associated with hierarchical probe content
jointly guides view placement, while the exploration
Context response determines the surrounding context
to retain.
Its verification probe describes evidence needed
to assess emerging hypotheses, turning verification
into guidance for constructing a new observation.
Final coordinate predictions are integrated through
uncertainty-weighted spatial voting.
This design connects semantic evidence descriptions
to adaptive visual inspection, complementing search
strategies guided by coordinate uncertainty and
candidate verification scores.

\section{Implementation Details}
All experiments were conducted on a single NVIDIA RTX PRO 6000 GPU. We set the history coverage discount $\rho$ to $0.5$ for 7B/8B models and $0.8$ for the 32B model. The candidate count $K$ and standard window dimensions $(w_0,h_0)$ follow the default settings of MVP~\cite{mvp} and are identical to those used in our MVP reproduction. We developed SPARK using Qwen3-VL-8B on a 300-example development subset of ScreenSpot-Pro. We selected the 7B/8B hyperparameters in this setting and used them without further tuning across all evaluated 7B/8B backbones and benchmarks. For Qwen3-VL-32B, we selected hyperparameters separately on the same development subset and held them fixed across all evaluated benchmarks.

\section{Detailed Prompt Templates}
\label{app:prompts}

SPARK uses two semantic probe templates and backbone-specific localization prompts.
The exploration and verification templates are shared across backbones.
Each user message contains the corresponding image followed by the original instruction.
In the prompt templates, \texttt{Coarse} and \texttt{Fine} are aliases for Context and Detail, respectively.

\begin{promptbox}{Exploration Probe}{promptBlue}
[System]
You are an expert UI element locator. Given a GUI image and a user's element description, identify the specified element using only visible evidence. The image resolution is height {height} and width {width}.
Describe the target from coarse to fine:
Coarse: name its likely containing panel or screen region and a visible landmark.
Fine: name the requested UI control and a visible detail distinguishing it from nearby controls.
Output only these two short phrases, at most 15 words per phrase. Do not output coordinates.

[User]
<Global screenshot>
{instruction}

[Assistant prefill]
Coarse:
\end{promptbox}

The Context and Detail content budgets are 24 and 40 tokens, respectively.

\begin{promptbox}{Verification Probe}{promptPurple}
[System]
Use visible evidence to identify the control requested by the original instruction. The marked locations are unverified hypotheses from previous observations. Do not assume a marked location is correct. Inspect the full screenshot and use another visible region if the hypotheses do not fit the requested function. Describe evidence needed to verify the requested control, not a final choice.
Output short phrases:
Control: <the requested control and identifying visible text or icon>
Context: <the containing panel or a relevant nearby visible landmark>
Inspect: <one specific visible detail needed to verify the target's function or exact location>
If no actionable evidence is visible, write Inspect: UNCLEAR.
Do not invent details. Do not output a winner, a label choice, or coordinates. In Inspect, name image evidence rather than the drawn marks or their labels.

[User]
<Global screenshot with hypothesis markers>
{instruction}

[Assistant prefill]
Control:
\end{promptbox}

The Control, Context, and Inspect content budgets are
16, 24, and 32 tokens, respectively, with an overall
generation cap of 96 tokens.
The Context and Inspect responses provide spatial
guidance for the verification observation.

\begin{promptbox}{Localization: Qwen3-VL and UI-TARS}{promptOrange}
[System]
You are an expert UI element locator. Given a GUI image and a user's element description, provide the coordinates of the specified element as a single (x,y) point. The image resolution is height {height} and width {width}. For elements with area, return the center point.

Output the coordinate pair exactly:
(x,y)

[User]
<Current global or local view>
{instruction}
\end{promptbox}

The image dimensions are instantiated using the
actual input view. The same localization template is
used for global and local observations.

For Qwen2.5-VL, localization follows its native
\texttt{computer\_use} tool interface.
The user message contains the current view and the
original instruction, and generation continues from
the following assistant prefix.

\begin{promptbox}{Localization: Qwen2.5-VL Assistant Prefill}{promptOrange}
<tool_call>
{"name": "computer_use", "arguments": {"action": "left_click", "coordinate": [
\end{promptbox}

\section{Algorithm Workflow of SPARK}
\label{sec:detailed_algorithm}

Algorithm~\ref{alg:spark} summarizes the SPARK procedure. 

\newcommand{\algstage}[1]{%
    \Statex \colorbox{gray!15}{%
        \parbox{\dimexpr\linewidth-2\fboxsep\relax}{%
            \strut\textbf{#1}%
        }%
    }%
}

\begin{algorithm}[t]
\caption{SPARK}
\label{alg:spark}
\small
\begin{algorithmic}[1]
\Require Screenshot $I$, instruction $d$, frozen VLM
$f_\theta$, local-view budget $B \geq 1$,
visual token budget $N$
\Ensure Target coordinate $p^\star$

\algstage{Initial observations}
\State $R_0 \gets$ full screenshot region
\State $(p_0,\bar H_0,A_G)
    \gets \Call{Ground}{I,d,R_0}$
\State $\mathcal{D} \gets
    \{(p_0,\bar H_0)\mid p_0\text{ is valid}\}$;
    $\mathcal{H} \gets \emptyset$
\State $R_1 \gets \Call{Init}{A_G,w_0,h_0}$
\State \Call{Observe}{$R_1$}; $b \gets 1$

\algstage{Probe-guided observations}
\State $(C^E,D^E)
    \gets \Call{ExplorationProbe}{I,d}$
\State $\mathcal{P} \gets
    \Call{BuildCandidates}{C^E,D^E,A_G}$
\While{$b < B$}
    \State $R \gets
        \Call{Focus}{C^E,D^E,C^E,\mathcal{P},\mathcal{H}}$
    \State \Call{Observe}{$R$}; $b \gets b+1$
    \State $\mathcal{C} \gets
        \Call{ClusterAndScore}{\mathcal{D}}$
    \If{$b = 2$ \textbf{and} $b < B$}
        \State $P^{\mathrm{hyp}} \gets$ medoids of the top $\min(2, |\mathcal{C}|)$ clusters
        \State $I^{\mathrm{hyp}} \gets
            \Call{Mark}{I,P^{\mathrm{hyp}}}$
        \State $(C^V,I^V) \gets
            \Call{VerificationProbe}{I^{\mathrm{hyp}},d}$
        \State $R \gets
            \Call{Focus}{C^V,I^V,C^E,\mathcal{P},\mathcal{H}}$
        \State \Call{Observe}{$R$}; $b \gets b+1$
    \EndIf
\EndWhile

\algstage{Uncertainty-weighted aggregation}
\State $\mathcal{C} \gets
    \Call{ClusterAndScore}{\mathcal{D}}$
\State $C^\star \gets
    \arg\max_{C\in\mathcal{C}} S(C)$
\State \Return $p^\star \gets
    \Call{Medoid}{C^\star}$
\end{algorithmic}
\end{algorithm}

\noindent\textbf{Algorithm conventions.}
The probe routines return normalized content-token
attention maps.
\textsc{BuildCandidates} constructs a shared pool from
the top-$K$ locations in each input map.
\textsc{Focus} selects a window using history-discounted
contextual and local evidence, then adapts its extent
using the undiscounted exploration Context map $C^E$.
\textsc{Observe} grounds the selected crop within budget
$N$, appends valid predictions and their entropies to
$\mathcal{D}$, and updates $\mathcal{H}$.
\textsc{ClusterAndScore} performs complete-link clustering
and uncertainty-weighted scoring as defined in the main text.

\section{Results on the Larger Model}
\label{sec:large}

We further evaluate SPARK with Qwen3-VL-32B on
ScreenSpot-Pro, OSWorld-G, and UI-Vision.
Table~\ref{tab:qwen3-32b-benchmarks} reports the complete
results under our evaluation setting.
SPARK achieves accuracies of 75.9\%, 77.8\%, and 44.7\%,
respectively, improving over single-view inference
by 20.5, 6.2, and 8.5 pp.
It also exceeds MVP by 1.4, 2.1,
and 1.1 pp across these benchmarks.
These results extend the benefits of probe-guided
adaptive focusing to a larger backbone, showing that
semantic inspection guidance remains useful as
model capacity increases.

\section{Uncertainty-weighted Voting for MVP}
\label{sec:mvp-weighted-voting}

We apply uncertainty-weighted voting to MVP to examine
its effectiveness beyond SPARK's view construction.
As shown in Table~\ref{tab:mvp-weighted-voting},
this aggregation improves MVP across all three
backbones, reaching 57.1\%, 50.3\%, and 66.4\%
with UI-TARS, Qwen2.5-VL, and Qwen3-VL, respectively.
SPARK further achieves 58.2\%, 52.5\%, and 68.7\%.
These results demonstrate that uncertainty weighting
can improve existing multi-view predictions,
while SPARK's complete framework delivers additional
gains through probe-guided observation and aggregation.
\begin{table}[!t]
  \centering
  \caption{Uncertainty-weighted aggregation of MVP predictions
  on ScreenSpot-Pro. Accuracy (\%).}
  \label{tab:mvp-weighted-voting}
  \scriptsize
  \setlength{\tabcolsep}{4pt}
  \resizebox{\columnwidth}{!}{%
    \begin{tabular}{@{}l
      >{\columncolor{gray!5}}r
      r
      >{\columncolor{gray!5}}r@{}}
      \toprule
      Method
        & \UITARSIcon\ 1.5-7B
        & \QwenIcon\ 2.5-VL-7B
        & \QwenIcon\ 3-VL-8B \\
      \midrule
      MVP 
        & 55.9 & 49.1 & 65.2 \\
      
      MVP + Uncertainty-weighted voting
        & 57.1 & 50.3 & 66.4 \\

     \rowcolor{violet!15}
      \textbf{SPARK}
          & \textbf{58.2} & \textbf{52.5} & \textbf{68.7} \\
      \bottomrule
    \end{tabular}%
  }
\end{table}

\section{Comparison of Baseline Methods with More Time Budget}

As shown in Table~\ref{tab:mvp-view-budget}, SPARK's average inference time of 16.79 seconds falls between those of MVP with six and eight local views (13.75--17.53 seconds), while achieving 3.2--3.5 percentage points higher accuracy. Increasing MVP's view budget does not close this gap, consistent with its reported view-count ablation, where additional views yield limited gains and can even reduce accuracy~\cite{mvp}.

\begin{table}[!t]
  \centering
  \caption{Accuracy and average inference time on ScreenSpot-Pro
  with Qwen3-VL-8B under different local-view budgets.}
  \label{tab:mvp-view-budget}
  \scriptsize
  \setlength{\tabcolsep}{5pt}
  \renewcommand{\arraystretch}{1.08}
  \begin{tabularx}{\columnwidth}{@{}
    >{\raggedright\arraybackslash}p{0.22\columnwidth}
    >{\columncolor{gray!5}\centering\arraybackslash}X
    r
    >{\columncolor{gray!5}}r@{}}
    \toprule
    Method & Local views & Acc.$\uparrow$ (\%)
      & Avg. time (s) \\
    \midrule
    MVP & 4 & 65.2 & 8.39 \\
    MVP & 6 & 65.2 & 13.75 \\
    MVP & 8 & 65.5 & 17.53 \\
    \midrule
    \rowcolor{violet!15}
    \textbf{SPARK} & 4 & \textbf{68.7} & 16.79 \\
    \bottomrule
  \end{tabularx}
\end{table}

\begin{table}[t]
  \centering
  \caption{Grounding accuracy (\%) on the 300-example development subset of ScreenSpot-Pro, the remaining 1,281 examples, and the full benchmark.}
  \label{tab:sspro-tuning-split}
  \small
  \setlength{\tabcolsep}{3pt}
  \renewcommand{\arraystretch}{1.12}
  \begin{tabular}{@{}l >{\columncolor{gray!5}}c c
                       >{\columncolor{gray!5}}c@{}}
    \toprule
    \textbf{Model / Method}
      & \shortstack{Development\\subset}
      & \shortstack{Remaining\\examples}
      & \shortstack{Full\\benchmark} \\
    \midrule
    \QwenIcon\ Qwen3-VL-32B & 56.0 & 55.3 & 55.4 \\
    \quad +MVP              & 76.3 & 74.1 & 74.5 \\
    \rowcolor{violet!15}
    \quad +SPARK & \textbf{77.0} & \textbf{75.6} & \textbf{75.9} \\
    \midrule
    \QwenIcon\ Qwen3-VL-8B  & 55.7 & 54.6 & 54.8 \\
    \quad +MVP              & 66.3 & 64.9 & 65.2 \\
    \rowcolor{violet!15}
    \quad +SPARK & \textbf{70.0} & \textbf{68.4} & \textbf{68.7} \\
    \bottomrule
  \end{tabular}
\end{table}

\section{Performance on the Development Subset and Remaining Examples}

We developed SPARK and selected its 7B/8B hyperparameters using
Qwen3-VL-8B on a 300-example development subset of ScreenSpot-Pro.
We used the same settings for all evaluated 7B/8B backbones across all
benchmarks without further tuning. For Qwen3-VL-32B, we selected
hyperparameters separately on the same development subset and held them
fixed across all evaluated benchmarks.

Table~\ref{tab:sspro-tuning-split} reports accuracy on the development
subset, the remaining 1,281 examples, and the full ScreenSpot-Pro
benchmark. SPARK outperforms MVP on both subsets with both Qwen3-VL-8B
and Qwen3-VL-32B. Together with the results on other backbones and
benchmarks in the main paper, these findings suggest that the selected
hyperparameters transfer across benchmarks and backbones.

\section{Limitations and Future Work}
\label{sec:limitations}

SPARK improves GUI grounding through training-free, probe-guided observations. As with other multi-view grounding methods, processing additional views incurs an inference cost. 
Adaptive stopping could help allocate the observation budget according to the evidence needed for each instruction. 
The usefulness of semantic guidance also depends on the backbone's ability to describe relevant visual evidence and associate it with spatial attention.
Future work could explore more robust probe extraction and complementary grounding signals.

\end{document}